\documentclass[runningheads]{llncs}
\usepackage{tabularx}
\usepackage{eccv}

\usepackage{eccvabbrv}
\usepackage{graphicx}
\usepackage{booktabs}

\usepackage[accsupp]{axessibility}  % Improves PDF readability for those with disabilities.
\usepackage{adjustbox}

\usepackage{hyperref}

\usepackage{orcidlink}

\usepackage{wrapfig}

\begin{document}

\title{Contrastive ground-level image and remote sensing pre-training improves representation learning for natural world imagery} 

\titlerunning{CRISP: ContRastive Image-remote Sensing Pre-training}

\author{Andy V. Huynh\inst{1,2,5}\orcidlink{0009-0003-2892-706X} \and
Lauren E. Gillespie\inst{1,2,3,4}\orcidlink{0000-0003-2496-8035} \and
Jael Lopez-Saucedo\inst{2}\and
Claire Tang\inst{2}\and
Rohan Sikand\inst{2}\and
Moisés Expósito-Alonso\inst{2,2,3,4,5}\orcidlink{0000-0001-5711-0700}}

\authorrunning{A. V. Huynh, L. E. Gillespie, et al.}

\institute{
Co-first authors\and
    Stanford University, Stanford CA 94305, USA \and
    Carnegie Science, Washington DC 20015, USA \and
    University of California, Berkeley, Berkeley CA 94720, USA \and
    Howard Hughes Medical Institute,  Chevy Chase MD  20815, USA \\
\email{\{avhuynh,gillespl\}@cs.stanford.edu} }

% \institute{
%     Princeton University, Princeton NJ 08544, USA \and
%     Springer Heidelberg, Tiergartenstr.~17, 69121 Heidelberg, Germany
%     \email{lncs@springer.com}\\
%     \url{http://www.springer.com/gp/computer-science/lncs} \and
%     ABC Institute, Rupert-Karls-University Heidelberg, Heidelberg, Germany\\
%     \email{\{abc,lncs\}@uni-heidelberg.de}
% }

% \vspace{-0.7cm}
\maketitle

% \vspace{-0.7cm}
\begin{abstract}
Multimodal image-text contrastive learning has shown that joint representations can be learned across modalities. Here, we show how leveraging multiple views of image data with contrastive learning can improve downstream fine-grained classification performance for species recognition, even when one view is absent. We propose ContRastive Image-remote Sensing Pre-training (CRISP)---a new pre-training task for ground-level and aerial image representation learning of the natural world---and introduce Nature Multi-View (NMV), a dataset of natural world imagery including $>3$ million ground-level and aerial image pairs for over 6,000  plant taxa across the ecologically diverse state of California. The NMV dataset and accompanying material are available at \url{hf.co/datasets/andyvhuynh/NatureMultiView}.

  \keywords{Contrastive learning \and Self-supervised learning \and Remote sensing \and Environment}
\end{abstract}
% \vspace{-0.7cm}
% \vspace{-0.5cm}
\section{Introduction}
\label{sec:intro}
% \vspace{-0.4cm}

\begin{figure} %[t]
% \vspace{-2mm}
        \begin{center}
                \includegraphics[width=0.9\linewidth]{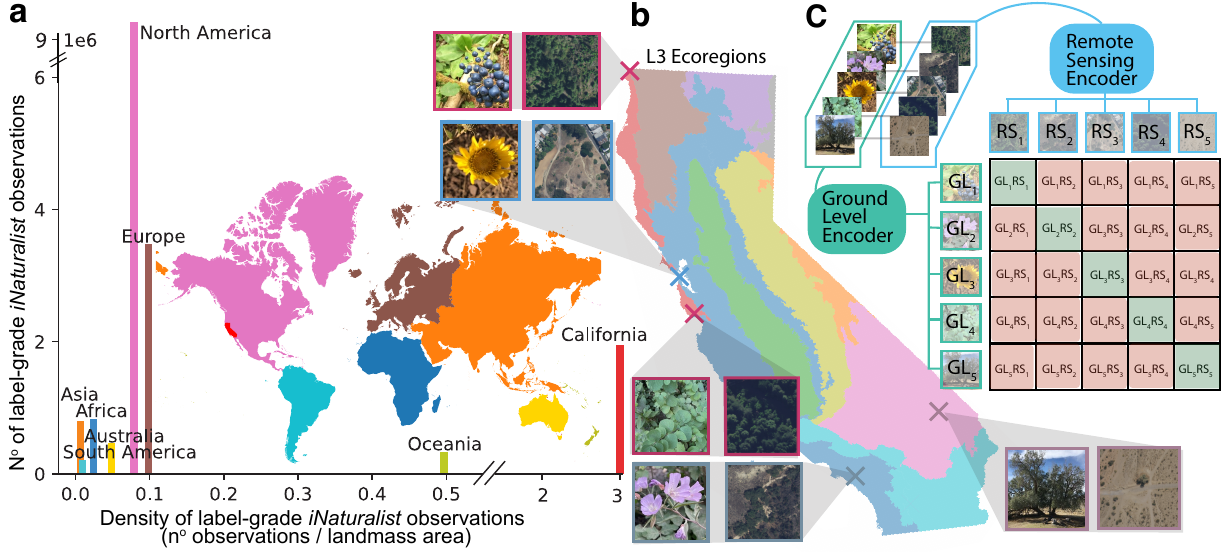}
        \end{center}
        % \vspace{-0.7cm}
% \vspace{-0.0cm}
\caption{\textbf{Overcoming label scarcity for natural world imagery.} \textbf{a.} Label inequity is a major problem for many publicly-available natural world imagery datasets, as many countries below the equator have relatively few label-quality observations per-unit area and biodiversity (e.g. South America, Africa). \textbf{b.} Images taken from the ground and from above at the same location often look visually similar and encode rich mutual information about natural world scenes and objects. \textbf{c.} The CRISP framework leverages this similarity to learn a joint representation between ground level-remote sensing image pairs so paired images from the same location (diagonal) have more similar representations than paired images from other locations (off-diagonal).
\label{fig:f1}
}
% \vspace{-1.7\baselineskip}
\end{figure}

In the natural world, there is a vast amount of unlabeled data. Users upload  millions of geo-tagged images of plants and animals to citizen science platforms like iNaturalist \cite{inat} yearly, and satellites collect remote sensing imagery from across the electromagnetic spectrum for most of the planet weekly. Yet, despite the size of these data sources, they often lack quality labels, limiting the use of fully-supervised machine learning methods with these data for critical downstream tasks like biodiversity monitoring.

Meanwhile, self-supervised methods---which utilize unlabeled data---have recently demonstrated comparable performance to fully-supervised methods across a wide variety of data modalities using much less labeled training data. While these methods have shown promising results for many vision tasks, these improvements have generally struggled to carry over to fine-grained classification tasks with natural world imagery, such as species recognition \cite{vanhorn2021benchmarking}. 

We argue that there is much information naturally shared between ground-level and aerial views of a scene which can be leveraged using self-supervised pre-training. Following this observation, we introduce ContRastive Image-remote Sensing Pre-training (CRISP), a self-supervised pre-training approach that compares and contrasts ground-level and aerial imagery to improve representation learning of natural world images.

To test our approach, we also introduce the new Nature Multi-View dataset which consists of $> 3$ million ground-level--aerial image pairs for over 6,000 plant taxa across California, an incredibly ecologically diverse state that has the highest diversity of any state in North America. This dataset provides a new benchmark for exploring how information shared between two different views can improve fine-grained classification tasks on natural world imagery.

 Compared to both standard Imagenet-pretrained models and self-supervised contrastive methods pre-trained with just one view in isolation, we find that CRISP multi-view pre-training improves downstream accuracy across a variety of class-balanced and unbalanced metrics for both species recognition and species distribution mapping. Encouragingly, the largest performance gains with CRISP come with the least amount of labeled data and the rarest classes, implying that multi-view approaches may be especially helpful in the data-sparse regions of the natural world that harbor the most biodiversity. CRISP further demonstrates improved few-shot transfer learning ability across aligned downstream natural world imagery tasks, including crop type mapping and urban tree species mapping. These findings highlight how multi-view imagery of the natural world contains rich information that can be useful even when few labels for training are present. % and how aligning representation learning frameworks to leverage this information frameworks can align with the underlying ecology how aligning representation learning frameworks with the underlying ecology can lead to marked performance improvements, demonstrating how domain knowledge can help adapt self-supervised representation learning methods that work for the human world to the natural world.

% \vspace{-0.5cm}
\section{Related Work}
\label{sec:related_work}
% \vspace{-0.2cm}
 
% \vspace{-0.2cm}
\subsection{Representation learning for fine-grained species recognition}
% \vspace{-0.3cm}
Fine-grained species recognition refers to the identification of a species—animal, plant, or other—from an image, often for biodiversity monitoring \cite{tuia2022perspectives}. Fully-supervised methods using large-scale biodiversity datasets have achieved impressive results \cite{garcin_plntnet-300k_2021}, but self-supervised methods often still struggle with heavily long-tailed, fine-grained biodiversity datasets \cite{vanhorn2021benchmarking}, leading to better performance mainly for well-represented classes \cite{vanhorn2017devil}. Even when leveraging additional unlabeled data from iNaturalist \cite{inat} for learning better representations for natural world imagery \cite{Cole_2022_CVPR, sagawa2022extending}, state-of-the-art contrastive frameworks still underperform on these fine-grained datasets, such as the iNat2021 dataset \cite{Van_Horn_2021_CVPR} where supervised methods still show strong performance \cite{Van_Horn_2021_CVPR, Cole_2022_CVPR}. Leveraging contextual information from image metadata---such as time and location---to mine informative positives for self-supervised learning from fixed-location camera trap imagery has shown some success over fully-supervised approaches \cite{Pantazis_2021_ICCV}, but these improvements have been limited to small datasets ($\leq 50$ species) with large charismatic animal species. 

Instead of utilizing image metadata to pick more informative triplet pairs, CRISP instead uses paired remote sensing imagery to improve representation learning for fine-grained species recognition. To this end, paired ground-level and remote sensing imagery datasets for species recognition exist, mainly the Auto Arborist \cite{beery2022auto} and BirdSAT \cite{sastry2024birdsat} datasets. While these datasets cover a larger geographic area than the Nature Multi-View dataset (continental North America and the world, respectively), importantly these datasets do not explore the many-to-one problem inherent in citizen science-based natural image collections (see Sec. \ref{sec:limitations}). These datasets also cover fewer species ($344$ and $1,486$, respectively vs. $6,602$ taxa in Nature Multi-View), while the Nature Multi-View dataset spans a much more diverse set of lifeforms, with species spanning 500 million years of evolutionary history, making it an upper bound for the challenge of fine-grained species recognition.

BirdSAT further proposed using masked autoencoders and the CLIP loss to learn joint representations between ground-level and aerial images of bird species \cite{sastry2024birdsat}. While BirdSAT's self-supervised framework also leverages the relationship between ground-level and remote sensing images using CLIP, in this work we additionally explore how breaking the symmetry between the ground-level and remote sensing similarity classification improves downstream performance, and further modify the CLIP loss to explicitly recognize when multiple ground-level images map to one aerial image in-batch \cite{graft-24}, which has interesting effects on downstream performance (Table \ref{tbl:species_recog}, S.M. Table \ref{tbl:clustering}).

% \vspace{-0.6cm}
\subsection{Representation learning for remote sensing imagery}
% \vspace{-0.4cm}
Remote sensing imagery---which includes aerial imagery---is widely used for many tasks, such as species distribution mapping \cite{botella2023overview, teng2023satbird, gillespie_image_2022}, land/crop type mapping \cite{jean2019tile2vec, manas_seasonal_2021, cong2022satmae}, semantic segmentation \cite{Mall_2023_CVPR},  and geo-localization \cite{cepeda2023geoclip, Haas_2024_CVPR}. Many of these approaches are self-supervised, leveraging the spatial \cite{jean2019tile2vec, 9140372}, temporal \cite{cong2022satmae, manas_seasonal_2021, Mall_2023_CVPR}, and/or sensor \cite{DBLP:journals/corr/abs-2108-05094} relationships between remote sensing images or bands. An especially powerful recent approach has been to exploit images' geolocations directly \cite{DBLP:journals/corr/abs-2011-09980, mai2023csp, cepeda2023geoclip, Haas_2024_CVPR}, or indirectly via associated variables such as land-cover \cite{li2021geographical} and more \cite{Bastani_2023_ICCV}.  

In this context, CRISP can be seen as another way of indirectly exploiting an image's geolocation for remote sensing representation learning by pairing ground-level images with remote sensing images, which has been a successful approach for vision-language remote sensing tasks \cite{graft-24} and wide-view image geolocalization \cite{Shugaev_2024_WACV,shi2019spatial, Liu_2019_CVPR, hu2018cvm, Cai_2019_ICCV,workman2015wide}. While CRISP pre-training is similar to wide-view image geolocalization in some aspects, the ultimate goals and challenges are complementary. Specifically, the goal of this work is to learn representations for fine-grained classification in long-tailed domains from little data and where performance on the rarest classes is exceedingly important. Therefore, the main challenges facing image geolocalization---including limited field-of-view \cite{Shugaev_2024_WACV} and inexact spatial co-registration of views \cite{Zhu_2021_CVPR}---are orthogonal to the goals of this work. Similarly image geolocalization benchmark datasets focus primarily on identifying the human world and are thus composed of scenes from the built environment (e.g. college campuses, suburban streets) \cite{vo2016localizing, Liu_2019_CVPR, Zhu_2021_CVPR, 10.1145/3394171.3413896} with relatively few classes and less focus on label imbalance \cite{workman2015wide}, meaning advances made on image geolocalization datasets---where the number of classes are usually fewer than a dozen---may not necessarily translate to improved rare species classification. 
%The complementary nature of these goals furthermore implies that progress in fine-grained species recognition from paired remote sensing + ground level imagery may
%image geolocalization and fine-grained species recognition 
%Therefore this work focuses on a complementary set of challenges and solutions compared to image geolocalization
%Nature Multi-View Dataset offers the first cross-view dataset specifically representing plant species in the natural world.

%%%%%%%%%%%%%%%%%%%%%%%%%%%%%%%%%%%%%%%%%%%%%%%%%%%%%%%%%%%%%%%%%%
% SECTION 3: DATASET 
%%%%%%%%%%%%%%%%%%%%%%%%%%%%%%%%%%%%%%%%%%%%%%%%%%%%%%%%%%%%%%%%%%

% \vspace{-0.4cm}
\section{The Nature Multi-View Dataset}
\label{sec:dataset}
% \vspace{-0.3cm}

\begin{figure}[t]
% \vspace{-2mm}
        \begin{center}
                \includegraphics[width=0.9\linewidth]{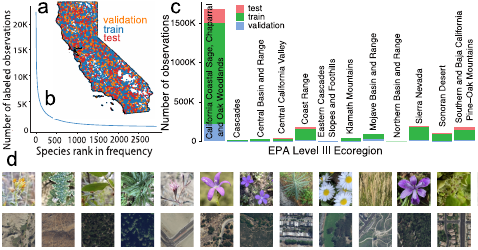}
        \end{center}
        % \vspace{-0.7cm}
        
    % \vspace{-0.0cm}
    \caption{\textbf{Overview of the Nature Multi-View dataset (NMV).} \textbf{a.} the 1,755,602 observations in the NMV dataset. \textbf{b.} Labels for identified observations in the NMV dataset exhibit a long-tail across classes, as is common in natural world settings. \textbf{c.} NMV observation density is also not uniform across space, a consequence of the opportunistic nature of citizen science datasets like iNaturalist \cite{inat}. \textbf{d.} A diverse set of ground level (top row) and paired aerial (bottom row) examples from the NMV dataset.
    \label{fig_f2}
    }
    
    % \vspace{-2.0\baselineskip}
\end{figure}

To encourage development of better machine learning methods for operating with diverse, unlabeled natural world imagery, we introduce Nature Multi-View (NMV), a multi-view ground-level and aerial image dataset for natural world imagery. The NMV dataset consists of over 3 million ground-level and aerial image pairs from over 1.75 million citizen science observations for over 6,000 native and introduced plant species across California (Table \ref{tbl:nmv}). We choose California as it is an extremely ecodiverse state which contains five of the world's ten major biomes (mountain, grassland, temperate forest, chaparral, desert), spans over a dozen unique ecoregions, and also contains some of the highest density of research-grade labeled images on Earth (Figure \ref{fig:f1}a), making it an ideal testbed for adapting self-supervised machine learning methods to natural world tasks.

%%%%%%%%%%%% SUBSECTION: curation %%%%%%%%%%%% 
% \vspace{-0.5cm}
\subsection{Nature Multi-View Dataset curation}
% \vspace{-0.3cm}
\textbf{Ground-level images.} First, ground-level observations and images were downloaded from the iNaturalist open data bucket on Amazon AWS \cite{inat}, filtering for images that were of vascular plants, that fell within the administrative bounds of the state of California, were taken between January 1, 2011 and September 27th, 2023, had a geographic uncertainty of less than 120 meters, were considered research-grade or in need of ID, and had remote sensing imagery available to pair. These ground-level photos are opportunistically-sampled, creative commons-licensed citizen science photos which span both natural and urban settings, and include observations from all of the major ecoregions of California (Figure \ref{fig_f2}a,d), meaning a wide variety of scene types and species are captured by the dataset (see S.M. \ref{inat_obs} for curation and S.M. \ref{sec:license} for license and reuse details).

\textbf{Aerial images.} For each of these ground-level images, we cropped a corresponding 256 by 256 pixel  60 cm-per-pixel resolution aerial-view RGB-Infrared image from the 2018 National Agriculture Imagery Program (NAIP) \cite{USDA_NAIP} centered on the latitude and longitude of the iNaturalist image (S.M. \ref{nmv_naip}). 

\begin{table}[h]
% \vspace{-0.7cm}
\footnotesize
\centering
\small
\caption{\textbf{Breakdown of Nature Multi-View (NMV) dataset by split.} The number of observations differs from the number of images as each observation may have multiple different ground-level images associated with it. Labels are the image's current community taxonomy assignment on iNaturalist (species, genus, etc.)
\label{tbl:nmv}}
\setlength\tabcolsep{2pt}
% \vspace{-0.1cm}
\begin{tabular}{lcccccccc}
\toprule
& \multicolumn{6}{c}{Train} & \multicolumn{1}{c}{Validation} & \multicolumn{1}{c}{Test} \\
\cmidrule(lr){1-1} \cmidrule(lr){2-7} \cmidrule(lr){8-8} \cmidrule(lr){9-9} 
  % Data split & Train & Train &  Train & Train & Train  &  Train  & Validation  & Test \\
Split \% ($\lambda$) & 100\% & 100\% & 20\% & 5\% & 1\% & 0.25\% & 100\% & 100\% \\
  Species labels? & $\times$ & $\checkmark$ & $\checkmark$ & $\checkmark$ & $\checkmark$ & $\checkmark$ & \textbf{\checkmark} & \textbf{\checkmark} \\
  \cmidrule(lr){1-1} \cmidrule(lr){2-7} \cmidrule(lr){8-8} \cmidrule(lr){9-9} 
% \midrule
Observations & 1,755,602 & 1,043,337 & 334,383  & 93,708 & 19,371 & 4,878 & 150,555 & 182,618 \\
Images & 3,307,025 & 1,954,542 & 390,908 & 97,727 & 19,545 & 4,886 & 279,114 &  334,887 \\
Classes & 6,602 & 2,946 & 2,914 & 2,767 & 2,205 & 1,421 & 2,946 & 2,946 \\
\bottomrule
\end{tabular}
% \vspace{-0.3\baselineskip}
% \vspace{-1.7\baselineskip}
% \\ The number of observations differs from the number of images as each observation may have multiple different ground-level images associated with it.
\end{table}
\textbf{Dataset splits.} To build the validation splits, we split California into a series of 0.1 degree by 0.1 degree boxes. We randomly designated 12.5\% of these boxes as test, 12.5\% as validation, and the remaining 75\% as training, and categorized all observations geographically located within said boxes as so. For the self-supervised partition of the NMV dataset, all observations were kept regardless of label quality, including images in need of ID (Table \ref{tbl:nmv} first column). For test and validation images, only research-grade observations ID'd down to the species level were kept (S.M. \ref{inat_crossval}). 

California has an unusually high density of well-labeled, research-grade iNaturalist compared to global averages (Figure \ref{fig:f1}a, S.M. Table \ref{tbl:nmv_density}), with almost 60\% of the 1.7 million observations in the NMV dataset considered research-grade. To better reflect realistic data availability in other biodiverse regions of the planet, for the fully supervised partition of the NMV dataset, we heavily restrict the size of the percentage splits ($\lambda$ in Table \ref{tbl:nmv}). For labeled observations, we provide $\lambda$s of 0.25\%---a similar relative size to publicly-available citizen science observations below the equator (e.g. Africa, and South America in S.M. Table \ref{tbl:nmv_density})---1\% and 5\%---similar in size to public citizen science observations above the equator (e.g. Europe, North America in S.M. Table \ref{tbl:nmv_density})---and 20\%---similar to the size of other natural world imagery datasets \cite{Horn_2018_CVPR, garcin_plntnet-300k_2021, vanhorn2021benchmarking} (S.M. \ref{inat_splits}).
%  These percentages roughly align with the expected labeled dataset size using average density of public citizen science observations in continents both below the equator (an estimated 0.302\%, 3.11\% 0.148\% and 0.012\% of the dataset for Australia, Oceania, Africa, and South America, respectively), and above (an estimated 0.113\%, 6.712\%, and 6.66\% of the dataset for Asia, Europe, and North America, respectively). We also include a split of 20\% which is similar to the size of other existing fully-labeled natural world imagery datasets \cite{Horn_2018_CVPR, garcin_plntnet-300k_2021}.

%%%%%%%%%%%% SUBSECTION: limitations %%%%%%%%%%%% 

% \vspace{-0.6cm}
\subsection{Dataset limitations}\label{sec:limitations}
% \vspace{-0.2cm}
\textbf{Geographic coverage.} The dataset only spans the state of California and is highly geographically heterogeneous as citizen science data is often geographically-skewed to more densely-populated and visited regions, such as major urban cities and national parks within the California Chaparral ecoregion (Figure \ref{fig_f2}c).

\textbf{Many-to-one pairing.} On average, the NMV dataset contains two ground-level images for every observation, meaning multiple views of the same plant are paired to the exact same aerial image (Table \ref{tbl:nmv}).

\textbf{Long-tail distribution.} Citizen science data often have long-tailed class imbalance, where common species are documented more often than rare species (Figure \ref{fig_f2}b). While this imbalance is a challenge for many machine learning methods, it is a realistic representation of the distribution biodiversity data \cite{enquist_commonness_2019}.

Taken together, these limitation make the NMV dataset a challenging but important benchmark for representation learning under various types of biases present in natural world imagery data.

%%%%%%%%%%%%%%%%%%%%%%%%%%%%%%%%%%%%%%%%%%%%%%%%%%%%%%%%%%%%%%%%%%
% SECTION 4: Methods
%%%%%%%%%%%%%%%%%%%%%%%%%%%%%%%%%%%%%%%%%%%%%%%%%%%%%%%%%%%%%%%%%%

% \vspace{-0.5cm}
\section{The CRISP Framework}
\label{sec:methods}
% \vspace{-0.3cm}

Here we overview the ContRastive Image-remote Sensing Pre-training (CRISP) framework, a contrastive approach for multi-view image-image pre-training inspired by CLIP \cite{radford2021learning}. Specifically, CRISP replaces the text embedding head of CLIP with another image encoder; thus given a batch of of $N^2$ image-image pairs, CRISP aims to predicts which ground-level and aerial image views are paired by learning a joint representation between both image views (Figure \ref{fig:f1}c) such that images from multiple views of the same location have a more similar representation than those from different locations\cite{oord_representation_2019}. We describe four different variations of CRISP in the following section. %Specific implementation and training details can be found in S.M. \ref{crisp}
% \end{itemize}

% \vspace{-0.0cm}
\subsection{Standard CRISP objective}
% \vspace{-0.0cm}
For standard CRISP---which mimics standard CLIP \cite{radford2021learning}---each batch consists of a two part loss: $L^{gl}$ which measures how aligned each ground-level representation is with the representations of the in-batch aerial images, and $L^{a}$ which does the same for the aerial image representation, as follows (see S.M. \ref{crisp} for details):

% l^{gl}_{i} = -\text{log} \frac{\text{exp}(\text{sim}(\boldsymbol{z}^{gl}_i, \boldsymbol{z}^a_i) / \tau)}{\sum_{k=1}^{N} \text{exp}(\text{sim}(\boldsymbol{z}^{gl}_i, \boldsymbol{z}^a_k) / \tau)}
\begin{equation}
    L^{gl} = \frac{1}{N} \sum_{i=1}^N -\text{log} \frac{\text{exp}(\text{sim}(\boldsymbol{z}^{gl}_i, \boldsymbol{z}^a_i) / \tau)}{\sum_{k=1}^{N} \text{exp}(\text{sim}(\boldsymbol{z}^{gl}_i, \boldsymbol{z}^a_k) / \tau)}
\label{eq:standard_gl}
\end{equation}

% l^{a}_{i} = -\text{log} \frac{\text{exp}(\text{sim}(\boldsymbol{z}^{gl}_i,\boldsymbol{z}^a_i)/ \tau)}{\sum_{k=1}^{N} \text{exp}(\text{sim}(\boldsymbol{z}^{gl}_k, \boldsymbol{z}^a_i)/ \tau) }
\begin{equation}
    L^{a} = \frac{1}{N} \sum_{i=1}^N -\text{log} \frac{\text{exp}(\text{sim}(\boldsymbol{z}^{gl}_i,\boldsymbol{z}^a_i)/ \tau)}{\sum_{k=1}^{N} \text{exp}(\text{sim}(\boldsymbol{z}^{gl}_k, \boldsymbol{z}^a_i)/ \tau) }
\label{eq:standard_a}
\end{equation}

where $\boldsymbol{z}^{gl}$ refers to a ground-level image representation, $\boldsymbol{z}^{a}$ refers to an aerial image representation, sim($\boldsymbol{u}, \boldsymbol{v}$) refers to the cosine similarity between two vectors $\boldsymbol{u}$ and $\boldsymbol{v}$, and $\tau$ is a temperature parameter. Standard CRISP uses a symmetric cross entropy loss, namely the unweighted mean of the two losses:

\begin{equation}
L = \frac{1}{2} ( {L^{gl} + L^a}  )
\label{eq:standard_final}
\end{equation}

%%%%%%%%%%%% SUBSECTION: Data augmentation %%%%%%%%%%%%

% \vspace{-0.8cm}
\subsection{CRISP objective with remote sensing data augmentation}
% \vspace{-0.3cm}

Within the NMV dataset, there are many ground-level images that map to the same aerial image. To overcome this many-to-one problem, we first explored augmenting the aerial imagery to provide a higher diversity of aerial views for the paired ground-level imagery. Specifically, we pre-trained CRISP with standard remote sensing cropping augmentations (CRISP-Aug), including random cropping, random horizontal and vertical flips, and random 90 degree rotations \cite{9140372, cong2022satmae} (see S.M. \ref{crisp_aug} for details).

%%%%%%%%%%%% SUBSECTION: M2o Loss %%%%%%%%%%%% 

% \vspace{-0.0cm}
\subsection{Many-to-one CRISP objective}
% \vspace{-0.0cm}

To better explicitly for the many-to-one nature of the NMV dataset in the representation learning process, we modified the CLIP loss similar to \cite{graft-24} and consider multiple positives within a single batch for each view, specifically including views that are not paired but are within a 250 meter radius of each other as positive. For this many-to-one version of CRISP loss (CRISP-M2o), Equations \ref{eq:standard_gl} and \ref{eq:standard_a} are updated as follows:

% \begin{equation}
%     \small L^{gl}=\frac{1}{N_a} \sum_{j=1}^{Na} \frac{1}{N_j}\sum_{i=1}^{N}\frac{1}{N_i}\sum_{j=1}^{N_i}-\log \frac{\exp(\text{sim}(\boldsymbol{z}^{gl}_{i,j},\boldsymbol{z}^a_i)/ \tau)}{\sum_{u=1}^{n} \sum_{a=1}^{N}\sum_{b=1}^{N_a} \exp(\text{sim}(\boldsymbol{z}^{gl}_i, \boldsymbol{z}^a_{a,b})/ \tau)}
% \end{equation}

% \begin{equation}
%     \small L^{a}=\frac{1}{N^GL}\sum_{i=1}^{N}\frac{1}{N_i}\sum_{j=1}^{N_i}-\log \frac{\exp(\text{sim}(\boldsymbol{z}^{gl}_{i,j},\boldsymbol{z}^a_i)/ \tau)}{\sum_{a=1}^{N}\sum_{b=1}^{N_a} \exp(\text{sim}(\boldsymbol{z}^{gl}_{a,b}, \boldsymbol{z}^a_i)/ \tau)}
% \end{equation}

\begin{equation}
    \small L^{gl}=\frac{1}{N^{gl}}\sum_{i=1}^{N^{gl}}\frac{1}{N_i^{a}}\sum_{j=1}^{N_i^{a}}-\log \frac{\exp(\text{sim}(\boldsymbol{z}^{gl}_{i},\boldsymbol{z}^a_{i,j})/ \tau)}{\sum_{m=1}^{N^{gl}}\sum_{n=1}^{N_m^{a}} \exp(\text{sim}(\boldsymbol{z}^{gl}_{i}, \boldsymbol{z}^a_{m,n}/ \tau)}
\label{eq:m2o_gl}
\end{equation}

\begin{equation}
    \small L^{a}=\frac{1}{N^a}\sum_{i=1}^{N^a}\frac{1}{N_i^{gl}}\sum_{j=1}^{N_i^{gl}}-\log \frac{\exp(\text{sim}(\boldsymbol{z}^{gl}_{i,j},\boldsymbol{z}^a_i)/ \tau)}{\sum_{m=1}^{N^a}\sum_{n=1}^{N_m^{gl}} \exp(\text{sim}(\boldsymbol{z}^{gl}_{m,n}, \boldsymbol{z}^a_i)/ \tau)}
\label{eq:m2o_a}
\end{equation}

In Equation \ref{eq:m2o_gl}, $N^{gl}$ is the number of ground-level images in a batch, while $N_i^{a}$ and $N_m^{a}$ are the number of aerial images for ground-level images $i$ and $m$, respectively. Symmetrically, in Equation \ref{eq:m2o_a}, $N^a$ is the number of aerial images in a batch, while $N_i^{gl}$ and $N_m^{gl}$ are the number of ground-level images for aerial images $i$ and $m$, respectively (see S.M. \ref{crisp_m20} for details).

%%%%%%%%%%%% SUBSECTION: Parametric CRISP objective %%%%%%%%%%%% 

\vspace{-0.0cm}
\subsection{Parameterized CRISP objective}
\vspace{-0.0cm}

To account for the fact that not all ground-level and aerial images may be equally informative during pre-training, we further introduce a parameterized version of the standard CRISP loss (CRISP-Par) that dynamically learns to weigh contributions from the two-part CRISP pre-training loss (Equation \ref{eq:standard_final}): 

\begin{equation}
    L = \sigma(\textbf{w}) L^{gl} + (1 - \sigma(\textbf{w})) L^{a},  
\label{eq:param}
\end{equation}

where $L^{gl}$ and $L^{a}$ refers to Equations \ref{eq:standard_gl} and \ref{eq:standard_a}, $\sigma$ is sigmoid, and $\textbf{w}$ is a learned parameter (see S.M. \ref{crisp_par} for details).

%  Specifically, we pre-trained CRISP both without and with image augmentations. Our different models include:

% % \vspace{-0.3cm}
% \begin{itemize}
%     \item \textbf{CRISP.} The standard CRISP framework trained without any data augmentation other than Imagenet and NAIP mean normalization for each respective view.
%     \item \textbf{CRISP-Aug.} CRISP with aerial view cropping augmentations that include random cropping, random horizontal flip, random vertical flip, and random rotation by 90 degree increments \cite{9140372}.
% \end{itemize}
% % \vspace{-0.3cm}

%%%%%%%%%%%%%%%%%%%%%%%%%%%%%%%%%%%%%%%%%%%%%%%%%%%%%%%%%%%%%%%%%%
% SECTION 5: Experiments
%%%%%%%%%%%%%%%%%%%%%%%%%%%%%%%%%%%%%%%%%%%%%%%%%%%%%%%%%%%%%%%%%%

% \vspace{-0.5cm}
\section{Experiments}
\label{sec:experiments}
% \vspace{-0.3cm}

\begin{figure}[t]
% \vspace{-2mm}
        \begin{center}
                \includegraphics[width=0.9\linewidth]{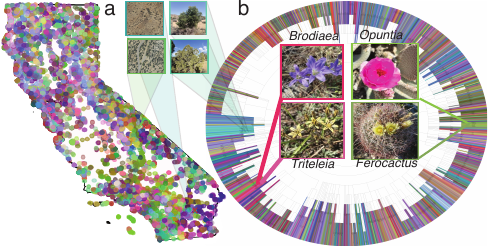}
        \end{center}
        % \vspace{-0.7cm}
        
% \vspace{-0.0cm}
\caption{\textbf{CRISP pre-trained representations  recapitulate ecological expectations.} \textbf{a.} UMAP projection of CRISP aerial view encoder embeddings in color space overlaid over map of aerial image locations. Aerial embeddings appear to be more similar for aerial images taken from similar ecosystems even if far away (see inset). \textbf{b.} UMAP projection of CRISP ground view encoder embeddings in color space overlaid over taxonomic tree of genera in dataset. Ground level embeddings appear to be more similar for more related genera, along with visually similar genera that may not be related (see inset).
\label{fig:f3}
}

% \vspace{-1.7\baselineskip}
\end{figure}

To determine the effectiveness of CRISP pre-training for representing natural world imagery, we consider four downstream fine-grained classification tasks: species recognition, species distribution modeling,  cropland mapping, and tree genus identification. In our work, we continue to focus specifically on plant species, although our approaches should be readily usable for other natural world tasks.
% \vspace{-0.4cm}
\subsection{Fine-grained species recognition}
% \vspace{-0.3cm}
In this work, we consider species recognition to be the fine-grained visual classification of what species is present in a ground-level image. For this task, we report Top-1 and Top-5 accuracy, and given the highly long-tailed nature of our data also report Top-1 macro-accuracy (accuracy averaged per-class) (Table \ref{tbl:species_recog}). To further disambiguate how the methods perform for common versus rare species, we also report Top-1 macro-accuracy for frequent, common, and rare species as defined in \cite{beery2022auto}, which we refer to as Top-1-F, Top-1-C and Top-1-R accuracy (S.M. Table \ref{tbl:t1_spec_recog_far}), along with Top-1 accuracy averaged per-ecoregion (S.M. Table \ref{tbl:t1_eco_accuracy}, see Figures \ref{fig:f1}b and \ref{fig_f2}c for ecoregion definitions). Accuracy metric and training details can be found in S.M. \ref{metss} and \ref{spec_rec}.
% \vspace{-0.6cm}

For species recognition, we fine-tune the pre-trained CRISP, CRISP-Aug, CRISP-M2o, and CRISP-Par ground image encoders (S.M. \ref{crisp}) along with the following supervised and self-supervised baselines: 
% \vspace{-0.2cm}
\begin{itemize}
    \item \textbf{Imagenet}, an Imagenet pre-trained ResNet50 \cite{DBLP:journals/corr/HeZRS15} (S.M. \ref{full_sup}),
    \item \textbf{MAE-NMV} a  vision transformer pre-trained on ground-level NMV imagery using the masked auto-encoding (MAE) self-supervised algorithm (S.M. \ref{mae}), 
    \item \textbf{MAE-Img} a vision transformer pre-trained on Imagenet imagery with MAE \cite{MaskedAutoencoders2021} (S.M. \ref{mae}), and
    \item \textbf{MoCo-v2}, a ResNet50 pre-trained on ground-level NMV imagery using the momentum contrast (MoCo-v2) contrastive algorithm \cite{chen2020improved} (S.M. \ref{moco}).
\end{itemize}

\begin{table}
% \vspace{-0.3cm}
\caption{\textbf{Species recognition accuracies.} Rows correspond to pre-trained backbones end-to-end fine-tuned. \label{tbl:species_recog}}
% \vspace{-0.2cm}
\centering
% \small % or \footnotesize
\begin{adjustbox}{width=\textwidth}
\begin{tabular}{lllllllllllll}
\toprule
& \multicolumn{4}{c}{Top-1 Acc} & \multicolumn{4}{c}{Top-5 Acc} & \multicolumn{4}{c}{Top-1 Macro Acc} \\
\cmidrule(lr){1-1} \cmidrule(lr){2-5} \cmidrule(lr){6-9} \cmidrule(lr){10-13}
Train \% ($\lambda$) & 0.25\% & 1\% & 5\% & 20\% & 0.25\% & 1\% & 5\% & 20\% & 0.25\% & 1\% & 5\% & 20\% \\
\cmidrule(lr){1-1} \cmidrule(lr){2-5} \cmidrule(lr){6-9} \cmidrule(lr){10-13}

CRISP & 36.05 & 50.55 & 61.24 & 70.74 & \textbf{51.66} & 68.28 & 78.56 & 86.28 & \textbf{11.36} & 20.86 & 28.81 & 39.90 \\
CRISP-Aug & 35.07 & 49.82 & 60.97 & 70.38 & 49.49 & 67.60 & 78.33 & 85.89 & 11.09 & 20.29 & 28.79 & 39.53 \\
CRISP-M2o & 35.06 & 49.43 & 60.47 & 70.77&50.68 &67.46 & 78.04& 86.26& 11.08& 20.26 & 28.27& 40.16\\
CRISP-Par & \textbf{36.18} & \textbf{51.04} & \textbf{61.61} &70.96 &51.51 & \textbf{69.38} & \textbf{79.16}& 86.56& 11.25 & \textbf{21.38} & 29.53& 40.31\\
\cmidrule(lr){1-1} \cmidrule(lr){2-5} \cmidrule(lr){6-9} \cmidrule(lr){10-13}
MAE-NMV & 6.34&22.64&46.70&65.45&15.87&41.44&67.84&83.07&0.91&4.70&16.68&34.19 \\
MAE-Img  & 5.00&27.03&55.24&73.45&13.05&46.61&74.58&88.08&0.61&6.08&24.83&45.32 \\
MoCo-v2 & 20.28 & 35.57 & 55.89 & 72.64 & 35.86 & 54.66 & 75.23 & 87.72 & 8.69 & 14.69  & 26.79 & 43.96 \\
Imagenet & 19.57 & 40.39 & 61.12 & \textbf{74.90} & 33.93 & 59.07 & 78.56 & \textbf{88.81} & 7.92 & 16.83 & \textbf{29.87} & \textbf{45.85} \\
\bottomrule
\end{tabular}
\end{adjustbox}
% \vspace{-2.0\baselineskip}
\end{table}

For species recognition from ground-level imagery---other than the largest $\lambda$=20\% split of the dataset---in general we see all CRISP objectives outperform all baselines by significant margins, including the Imagenet pre-trained backbone, which has previously shown strong performance on fine-grained species recognition tasks \cite{vanhorn2021benchmarking}. These improvements are most apparent in low data settings that mimic many real-world biodiversity dataset sizes (CRISP-Par avg. acc. improvement over baselines: 23.4\% Top-1 Accuracy @ $\lambda=0.25\%$, Table \ref{tbl:species_recog}), including when accounting for label imbalance (CRISP avg. acc. improvement over baselines: 6.8\% Top-1 Macro Accuracy @ $\lambda=0.25\%$, Table \ref{tbl:species_recog}) and geographic imbalance (CRISP-Aug avg. acc. improvement over baselines: 13.9\% Top-1 Eco-Accuracy  @ $\lambda=0.25\%$, S.M. Table \ref{tbl:t1_eco_accuracy}). 

Within the CRISP objectives, we see that CRISP-Par exhibits the strongest performance, possibly because it enables the learning process to focus on the most information-dense data view, which  for fine-grained species recognition is ground level-imagery. Interestingly, despite the fact that there are nearly two ground-level images for every aerial image in the NMV dataset, we see that the many-to-one objective in general underperforms when compared to both the standard CRISP and paramaterized CRISP objectives. Even when focusing on the observations with the most associated ground-level images, the many-to-one objective appears to be no better at clustering together ground-level image embeddings from the same location than the standard CRISP objective (S.M. \ref{embeddings}, S.M. Table \ref{tbl:clustering}). These results imply that CRISP pre-training can effectively learn which images are from the same local area even without explicitly receiving a signal for this relationship during training.

Similarly, projecting embeddings from the CRISP-pretrained ground-level encoder into color space with UMAP and mapping it to the NMV dataset's taxonomic hierarchy of genera (S.M. \ref{embeddings} Figure \ref{fig:f3}b) suggests the CRISP representation learning process has also captured some notion of the phylogenetic and 
physiological similarity between species. For example, genera \textit{Opuntia} and \textit{Ferocactus} are two closely related and succulent genera, and the CRISP embeddings correctly  cluster them as closer in color space than the less similar and less-related \textit{Brodiaea} and \textit{Triteleia}. This implies that CRISP pre-trained features could be used for not just species recognition, but other important ecological tasks, such as trait estimation from images.
%For the smallest dataset size at $\lambda=0.25\%$, we see CRISP pre-training increases downstream Top-1 classification accuracy by 16.48\% and 15.76\% and  Top-1 macro-accuracy by 3.43\% and 2.68\% over the fully supervised and contrastive baselines, respectively. 
% From these results, we see that CRISP pre-training is effective for species recognition in low-data scenarios over standard fully-supervised or contrastive baselines, specifically when the labeled dataset is extremely small ($\lambda=0.25\%$), reflective of the realized size of many biodiversity datasets (Figure \ref{fig:f1}). Furthermore, the fact that CRISP outperformed standard the contrastive approach trained using the same volume of ground level imagery implies that the information shared between different views of natural world imagery is useful and can be learned and exploited in a self-supervised manner. 
% Interestingly, these gains do not extend to the rarest species, implying there is still much work to be done to leverage the rich information shared between different views of natural world imagery.

% \vspace{-0.4cm}
\subsection{Species distribution mapping}

\begin{table*}
% \vspace{-0.7cm}
\small
\centering
\footnotesize
\caption{\textbf{Species distribution mapping accuracy.} Rows correspond to pre-trained backbones end-to-end fine-tuned. \label{tbl:sdm}}
% \vspace{-0.3cm}
\begin{tabular}{lllllllllllll}
\toprule
& \multicolumn{4}{c}{Top-5 Acc} & \multicolumn{4}{c}{Top-5 Macro Acc} & \multicolumn{4}{c}{Top-30 Macro Acc} \\
\cmidrule(lr){1-1} \cmidrule(lr){2-5} \cmidrule(lr){6-9} \cmidrule(lr){10-13}
Train \% ($\lambda$) & 0.25\% & 1\% & 5\% & 20\% & 0.25\% & 1\% & 5\% & 20\% & 0.25\% & 1\% & 5\% & 20\%\\
\cmidrule(lr){1-1} \cmidrule(lr){2-5} \cmidrule(lr){6-9} \cmidrule(lr){10-13}

CRISP & 8.30 & 7.36 & 6.52 & 6.91 & 3.48 & 3.70 & \textbf{4.02} & \textbf{4.82} & 13.53 & 14.29 & 14.89 & 17.37 \\
CRISP-Aug & 7.28 & 7.36 & 8.02 & 11.60 & 3.55 &	3.36 &	3.72 & 4.32 & 13.56 & 13.46 & 14.25 & 17.32 \\
CRISP-M2o & 8.53 &7.56 &6.45 & 7.05& \textbf{3.83}&3.78 & 3.87& 4.53&  14.09 &14.70 & 14.56& \textbf{17.42} \\
CRISP-Par & \textbf{9.30} & 7.92 & 6.73& 6.44& 3.61& \textbf{3.98}& 4.00&4.51 & \textbf{15.51} & \textbf{14.88} & \textbf{15.50}&17.16 \\
\cmidrule(lr){1-1} \cmidrule(lr){2-5} \cmidrule(lr){6-9} \cmidrule(lr){10-13}
SatMAE-NMV & 4.81& 9.50& \textbf{13.37}&\textbf{14.76}& 0.45& 1.21& 2.90& 4.62& 2.60& 5.09& 12.17& 16.47 \\
SatMAE-FMoW & 7.34	& 8.76& 	9.12& 		9.44& 1.32& 	1.89& 	1.81& 	3.31	&6.35	& 8.85	& 8.39& 	13.68\\
SauMoCo & 8.30 & \textbf{9.96} & 12.13 & 11.83 & 1.77 & 2.31& 3.02 & 3.82 & 8.12& 10.17 & 13.04& 15.53 \\
Random Init. & 4.98 & 8.36 & 10.90 & 11.26 & 0.48 & 1.23 & 2.08& 3.10 & 2.80 & 6.05 & 10.02 & 13.72 \\
\bottomrule
\end{tabular}
% \vspace{-2.0\baselineskip}
\end{table*}

Species distribution mapping (SDM)---defined here as predicting what species of plant is most likely to be present in an aerial image---is another fine-grained visual task that can be performed when only one view of the NMV dataset's natural world imagery is present. While under this formulation, SDM is essentially species classification from a different view, predicting species presence from remote sensing data is a critical \cite{randin_monitoring_2020} but challenging \cite{deneu_participation_2020} task for biodiversity monitoring. Given that multiple species may be present in the same aerial image, we report Top-5 Accuracy,  Top-5 and Top-30 Macro Accuracies (Table \ref{tbl:sdm}), Top-1 and Top-5 Accuracies, Top-1 Macro Accuracy (S.M. Table \ref{tbl:t1_sdm_acc}), accuracy broken down by class frequencies into frequent, common, and rare species (Table \ref{tbl:sdm_far}), and Top-1 accuracy averaged per-ecoregion (S.M. Table \ref{tbl:t1_eco_accuracy}), with implementation details found in S.M. \ref{sdm_exp}.

For species distribution mapping, we fine-tuned the pre-trained CRISP, CRISP-Aug, CRISP-M2o, and CRISP-Par aerial image encoders (S.M. \ref{crisp}) along with the following supervised and self-supervised baselines: 
% \vspace{-0.2cm}
\begin{itemize}
    \item \textbf{Random-Initialized Backbone}, a randomly-initialized ResNet50 \cite{DBLP:journals/corr/HeZRS15} (S.M. \ref{full_sup}), 
    \item \textbf{SatMAE-NMV}, a  vision transformer pre-trained on aerial NMV imagery using MAE (S.M. \ref{mae}),
    \item \textbf{SatMAE-FMoW}, a vision transformer pre-trained on Functional Map of the World (FMoW) imagery with MAE \cite{cong2022satmae} (S.M. \ref{mae}), and
    \item \textbf{SauMoCo}, a ResNet50 pre-trained on aerial NMV imagery using the augmentations from SauMoCo \cite{9140372} (S.M. \ref{moco}).
    
\end{itemize}

\begin{table*}
% \vspace{-0.5cm}
% \small
\centering
\caption{\textbf{Species distribution mapping broken down by species frequency.} Rows correspond to pre-trained backbones end-to-end fine-tuned. \label{tbl:sdm_far}}
% \vspace{-0.3cm}
\begin{adjustbox}{width=\textwidth}
\begin{tabular}{lllllllllllll}
\toprule
& \multicolumn{4}{c}{Top-30-F} & \multicolumn{4}{c}{Top-30-C} & \multicolumn{4}{c}{Top-30-R} \\
\cmidrule(lr){1-1} \cmidrule(lr){2-5} \cmidrule(lr){6-9} \cmidrule(lr){10-13}
Train \% ($\lambda$) & 0.25\% & 1\% & 5\% & 20\% & 0.25\% & 1\% & 5\% & 20\% & 0.25\% & 1\% & 5\% & 20\%\\
\cmidrule(lr){1-1} \cmidrule(lr){2-5} \cmidrule(lr){6-9} \cmidrule(lr){10-13}
CRISP & 18.99 & 23.78 &	20.26 &	20.28&	7.08&11.14&	18.72&	22.38 &	5.75 &	3.97 &	\textbf{7.22} &	\textbf{11.60} \\
CRISP-Aug & 17.67 & 21.42&	24.19&	33.37&	\textbf{8.84}&	10.64&	14.01&	16.94&	7.41&	5.01&	5.83&	6.13 \\
CRISP-M2o & 19.32&23.48&20.57&21.93&7.60&\textbf{11.45}&17.63&\textbf{23.16}&\textbf{7.50}&\textbf{5.63}&6.97&9.95\\
CRISP-Par & \textbf{21.96}&\textbf{25.01}& 23.23& 	19.49& 7.76& 11.21& \textbf{18.89}& 22.68& 6.67& 4.28& 6.20& 11.42\\
\cmidrule(lr){1-1} \cmidrule(lr){2-5} \cmidrule(lr){6-9} \cmidrule(lr){10-13}
SatMAE-NMV & 4.68&12.87&\textbf{32.65}&\textbf{37.87}&0.00&0.09&5.45&14.86&0.00&0.00&0.60&1.96\\
SatMAE-FMoW &10.92&19.48&22.72&27.6 &	0.90&3.07&3.46&13.61&	0.00&	0.42&	0.48&	3.52\\
SauMoCo &  13.49&	22.61& 30.94 &	32.72&	1.68&	3.57&	8.69&	14.80&	0.79&	0.08&	3.47&	1.68\\
Random Init. & 4.97&	14.51&	26.61&	30.89&	0.04&	0.92&	4.36&	12.27&	0.28&	0.07&	2.19&	0.85 \\
\bottomrule
\end{tabular}
\end{adjustbox}
\vspace{-1.0\baselineskip}
\end{table*}

% \vspace{-0.4cm}
For species distribution modeling from aerial imagery, CRISP pre-training exhibits a lower Top-5 Accuracy than the self-supervised baselines for all but the smallest $\lambda=0.25\%$ (Table \ref{tbl:sdm}). However, for accuracy metrics that correct for class imbalance, CRISP pre-training improves Top-5 and Top-30 Macro Accuracy by a meaningful margin (CRISP-Par avg. acc. improvement over baselines: 2.8\% for Top-5 Macro Accuracy; 10.5\% for Top-30 Macro Accuracy @ $\lambda=0.25\%$, Table \ref{tbl:sdm}). Breaking Top-30 Macro Accuracy down by species rarity, CRISP pre-training maintains predictive power for both the most common and the rarest classes (average CRISP pre-training accuracy @ $\lambda=0.25\%$: Top-30-F: 19.5\%, Top-30-R: 6.8\%, Table \ref{tbl:sdm_far}), unlike baseline approaches (average baseline accuracy @ $\lambda=0.25\%$: Top-30-F: 8.5\%, Top-30-R accuracy: 0.27\%, Table \ref{tbl:sdm_far}), a trend further reflected in accuracy broken down by ecoregion (CRISP-Aug avg. acc. improvement over baselines: 1.6\% @ $\lambda=0.25\%$ S.M. Table \ref{tbl:t1_eco_accuracy}).

\begin{table}[!b]
{
% \vspace{-0.2cm}
\footnotesize
\centering
\setlength\tabcolsep{2pt}
\caption{\textbf{Mixture of experts results on NMV aerial + ground imagery.} }
\begin{tabular}{rccccccccccccc}
% {lllllllllllll}
\toprule
& \multicolumn{4}{c}{Top-1 Acc} & \multicolumn{4}{c}{Top-1 Macro Acc} & \multicolumn{4}{c}{Top-1 Eco.} \\
\cmidrule(lr){1-1} \cmidrule(lr){2-5} \cmidrule(lr){6-9} \cmidrule(lr){10-13}
Train \% ($\lambda$) & .25\% & 1\% & 5\% & 20\% & .25\% & 1\% & 5\% & 20\% & .25\% & 1\% & 5\% & 20\% \\
\cmidrule(lr){1-1} \cmidrule(lr){2-5} \cmidrule(lr){6-9} \cmidrule(lr){10-13} 

CRISP &\textbf{12.4}&	\textbf{40.9}&	\textbf{61.8}&	73.0&	\textbf{3.2}&	14.5&	29.5&	43.3&	\textbf{6.7}&	\textbf{27.2}&\textbf{49.7} &	64.0	 \\
 Supervised & 4.9&	39.8&	61.3&	\textbf{74.7} &	1.3&	\textbf{16.4}&	\textbf{30.6}&	\textbf{47.4}&2.7&	27.0&	47.8&\textbf{66.3} \\
\bottomrule % TODO: reinsert?

\end{tabular}
% \vspace{-0.3cm}
\label{tbl:moe}
}
% \vspace{-0.3\baselineskip}
\end{table}
Beyond accuracy improvements, projecting embeddings from the CRISP-pretrained aerial encoder into color space with UMAP and mapping them across the landscape of California (S.M. \ref{embeddings}, Fig. \ref{fig:f3}a) reveals spatial patterns that imply CRISP pre-training captures a notion of environmental similarity even without an explicit training signal, similar to phylogenetic similarity for the ground-level encoder. For example, observations from Mojave Desert (inset) have more similar embeddings in colorspace, even if geographically separated by large distances. Again, the parameterized CRISP objective exhibits on average a stronger performance than the many-to-one  objective, similar to the ground-level view results.

It should be noted that in general all models have significantly worse performance predicting species labels from the aerial view than from ground-level view. Even when combining both encoders together in a mixture-of-experts (MoE) setting (S.M. \ref{moe}), in Table \ref{tbl:moe} and S.M. Table \ref{tbl:t5_moe}, performance still decreases when also using the aerial NMV imagery for species-level classification as compared to just the ground-level imagery (Top-1 Acc: 24.1\% decrease from Table \ref{tbl:species_recog}, Top-1 Macro Acc: 8.16\% decrease from Table \ref{tbl:species_recog}, Top-1 Eco: 14.3\% decrease, from S.M. Table \ref{tbl:t1_eco_accuracy}). That being said, it is generally known that species-level classification from aerial imagery alone is an exceedingly challenging and still unsolved task \cite{lacoste_geo-bench_2023}, and while there is yet much room for improvement for species classification from aerial imagery, it is promising that CRISP does significantly improve over the baseline approaches, especially in low-data settings for rare species.

% \vspace{-0.5cm}
\subsection{Crop type mapping}
% \vspace{-0.2cm}

Given the inherent challenges of predicting species labels from aerial imagery \cite{lacoste_geo-bench_2023}, we also compare CRISP's ability to generate fine-scaled crop type maps. Crop type mapping involves predicting the crop type of a given area from remote sensing imagery, a coarser-grained task than predicting species labels, but nevertheless a long-tailed and challenging task \cite{jean2019tile2vec}. Following Tile2Vec \cite{jean2019tile2vec}, we extracted NAIP aerial imagery features from frozen encoders, which were then used to fit a variety of downstream classifiers to predict 66 different crop classes from the 2016 Cropland Data Layer (CDL) in Central Valley, California (S.M. \ref{crop_mapping}). We also considered the following baselines:
% \vspace{-0.3cm}
\begin{itemize}
    \item \textbf{Tile2Vec}, a contrastive framework for aerial imagery using triplet loss and geographic distance to choose positive and negative images \cite{jean2019tile2vec} (S.M. \ref{crop_mapping}), 
    \item \textbf{SauMoCo}, as previously defined, and
    \item \textbf{Random}, where random values are sampled from a normal distribution.
\end{itemize}

\begin{table*}
% \vspace{-0.4cm}
\small
\centering
\caption{\textbf{Cropland mapping accuracies.} The crop type classification accuracy of different feature extraction methods and models (Logistic Regression, Multi-Layer Perceptron, Random Forest) with 1,000 and 10,000 training examples.\label{tbl:crop_map}}
% \vspace{-0.2cm}
\begin{tabular}{lcccccc}
\toprule
& \multicolumn{3}{c}{10,000 training examples} & \multicolumn{3}{c}{1,000 training examples} \\
 \cmidrule(lr){2-4} \cmidrule(lr){5-7} 
 &     LR &    MLP &     RF &     LR &    MLP &     RF \\
\cmidrule(lr){2-4} \cmidrule(lr){5-7}
CRISP &  \textbf{67.75} &  61.05 &  64.00 &  60.87 &  \textbf{61.26} &  59.18 \\
CRISP-Aug &  65.01 &  61.02 &  64.14 &  59.90 &  60.20 &  59.64 \\
CRISP-Par & 65.20 &  60.26 &  64.60 & 60.62& 60.19& 59.20  \\
CRISP-M2o & 66.20 &  61.11 & 64.37  & 57.54 & 58.57&  60.32 \\
%CRISP-M2o-Param &  &   &   & & &   \\
\cmidrule(lr){1-1}\cmidrule(lr){2-4}\cmidrule(lr){5-7}   
% SatMAE-NMV &  &   &   & & &   \\
% SatMAE \cite{cong2022satmae} &  &   &   & & &   \\
SauMoCo   &  63.90 &  63.38 &  64.07 &  60.07 &  59.59 &  58.03 \\
Tile2Vec  & 57.90 & 53.35 & 56.10 &53.55 & 50.65 &52.70 \\
Random   &  41.90 &  25.35 &  19.80 &  42.00 &  24.75 &  25.50 \\
\bottomrule
\end{tabular}
\vspace{-.7\baselineskip}
\end{table*}

CRISP outperform all three baselines both when using few examples ($1.2\%$ improvement with the multilayered perceptron classifier @ 1,000 examples) and many ($3.9\%$ improvement with logistic regression classifier @ 10,000 examples, Table \ref{tbl:crop_map}). Interestingly, CRISP-M2o performed worse with fewer training examples (1,000 examples, random forest) than with more (10,000 examples, logistic regression classifier). These results imply that the CRISP pre-training framework can yield useful representations even without fine-tuning, and that including ground-level data during pre-training improves performance for remote sensing tasks. 

% \vspace{-0.6cm}
\subsection{Tree genera identification with Auto Arborist}
% \vspace{-0.3cm}

To see how well CRISP pre-training extends to larger geographical areas, we also ran CRISP on the Auto Arborist (AA) dataset, a continent-scale dataset of over 3 million ground-aerial pairs of 328 urban tree genera across 23 cities in the U.S. and Canada\cite{beery2022auto}. Similar to the NMV dataset, we compare three types of model backbone pre-training on both ground-level and aerial views  using the entire AA training dataset (training details can be found in S.M. \ref{aa}):
% \vspace{-0.2cm}
\begin{itemize}
    \item aerial + ground-level \textbf{CRISP} with no aerial augmentations,
    \item aerial \textbf{SauMoCo} and separately ground-level \textbf{MoCo-v2},
  \item aerial \textbf{Random-Initialized} and separately ground-level \textbf{Imagenet}-pretrained
\end{itemize}
  % \vspace{-0.2cm}
 Towards our goal of improving fine-grained classification performance in low-data settings, these models were fine-tuned for one low-data but high diversity city from each region (East: Charlottesville, Central: Sioux Falls, West: Cupertino, see \cite{beery2022auto} Table 2 for city details). To further test extrapolation in low-data settings, the city-specific model's performance was calculated using the holdout city from that city's region  (East: Pittsburgh, Central: Boulder, West: Santa Monica).
% \vspace{-0.2cm}
% \begin{itemize}
%     \item aerial\ \textbf{Random} backbone ResNet50,
%     \item and ground-level \textbf{Imagenet} backbone ResNet50
%   % \vspace{-0.2cm}
% \end{itemize}

\begin{table*}
% \vspace{-0.4cm}
\small
\centering
\caption{\textbf{Top-5 Macro Accuracy on Auto Arborist Dataset.} We report accuracies for models fine-tuned on either aerial or ground level imagery.\label{tbl:aa}}
% \vspace{-0.2cm}
\begin{tabular}{lcccccc}
\toprule
& \multicolumn{3}{c}{Ground Level} & \multicolumn{3}{c}{Aerial} \\
\cmidrule(lr){2-4} \cmidrule(lr){5-7} 
 Test City &     Santa Monica &    Boulder &     Pittsburgh &     Santa Monica &    Boulder & Pittsburgh \\
 % \cmidrule(lr){2-4} \cmidrule(lr){5-7} 
 % Train City &     Cupertino &    Sioux Falls &     Charlottesville &   Cupertino &    Sioux Falls &     Charlottesville \\
\cmidrule(lr){2-4} \cmidrule(lr){5-7}
CRISP & \textbf{22.07} &	\textbf{17.88} &	\textbf{14.95} &	\textbf{13.02} &	\textbf{12.40} & \textbf{12.80} \\
MoCo-v2 / SauMoCo   & 13.97 &14.23 &	12.82 & 9.84 & 10.27 &	8.67 \\
Imagenet / Random   & 16.33 & 12.23 &	12.67 & 7.19 & 9.33 &	8.50 \\
% CRISP & \textbf{22.07} &	\textbf{17.88} &	\textbf{14.95} &	\textbf{13.02} &	\textbf{12.40} & \textbf{12.80} \\
% MoCo-v2   & 13.97 &14.23 &	12.82 & - &- & - \\
% SauMoCo   &  - & - & - & 9.84 & 10.27 &	8.67 \\
% Imagenet   & 16.33 & 12.23 &	12.67 & - & - & - \\
% Random   & - & - & - & 7.19 & 9.33 &	8.50 \\
\bottomrule
\end{tabular}
% \vspace{-1.7\baselineskip}
\end{table*}

% \vspace{-0.3cm}
Reflecting the results from the NMV dataset, CRISP outperforms all compared approaches for both views of data (avg. acc. increase: 4.6\% for ground-level, 3.25\% for aerial images, Table \ref{tbl:aa}), and Top-5 Accuracy as well (S.M. Table \ref{tbl:aa_t1_macro}). These results show that CRISP pre-training can improve representation learning performance for natural world imagery at continental-scale.

% \vspace{-0.4cm}
\section{Discussion}
\label{sec:discussion}
% \vspace{-0.3cm}
% We propose ContRastive Image-remote Sensing Pre-training (CRISP), a self-supervised contrastive framework to learn a shared representation space between ground-level image-aerial image pairs for a variety of downstream natural world classification tasks. We furthermore present the new Nature Multi-View (NMV) dataset, consisting of over 3 million ground-level image-aerial image pairs for over 6,000 species, standing as a new benchmark for multi-view self-supervised learning of the natural world. 

We propose ContRastive Image-remote Sensing Pre-training (CRISP), a self-supervised framework for natural world tasks, and present the Nature Multi-View (NMV) dataset, a new benchmark of over 3 million ground-level image-aerial image pairs. Multi-view pre-training on NMV with CRISP improves downstream per-species classification accuracy from both ground-level and aerial views over supervised and self-supervised single-view approaches, especially in low data settings that mirror biodiversity data availability of the world's most biodiverse yet most threatened regions. Further improvements for crop mapping (Table \ref{tbl:crop_map}) and tree prediction (Table \ref{tbl:aa}) suggest that the CRISP approach can learn broadly useful representations of the natural world and beyond, including for other multi-view tasks such as directing self-driving cars \cite{vo2016localizing} and forest monitoring \cite{ouaknine_openforest_2023}.
% CRISP further outperformed self-supervised single-view  approaches trained on the same volume of ground-level imagery,  implying the information shared between different views of natural world imagery can be effectively exploited in a self-supervised manner. Beyond species classification, CRISP's learned features capture known ecological (Figure \ref{fig:f3}a) and physiological structure (Figure \ref{fig:f3}b), implying that the CRISP pre-training process will useful for a variety of downstream ecological mapping tasks, such as crop type mapping (Table \ref{tbl:crop_map}) and urban tree species mapping (Table \ref{tbl:aa}). 
\newpage
\section*{Acknowledgements}
We thank Noah Goodman and Gabriel Poesia for their comments and discussion, and we further thank CoCoLab for donating compute for this project. We also thank the TomKat Center for Sustainable Energy for early feedback and discussion. We further thank Carnegie Science for funding support for this research (A.V.H, L.E.G., M.E.-A.). This research was also funded by the NSF Graduate Research Fellowship DGE-1656518 (L.E.G.), the TomKat Graduate Fellowship for Translational Research (L.E.G.), the Office of the Director of the National Institutes of Health's Early Investigator Award (1DP5OD029506-01, M.E.-A.), the Howard Hughes Medical Institue, and the University of California, Berkeley. Compute for this project was performed on the Calc cluster at Carnegie Science, the Caltech Resnick High Performance Computing Center, and the Stanford SC Compute Cluster. 


\clearpage
\setcounter{page}{1}

% \maketitlesupplementary

\title{Supplemental Materials for: Contrastive ground-level image and remote sensing pre-training improves representation learning for natural world imagery} 

\titlerunning{CRISP: ContRastive Image-remote Sensing Pre-training}

\author{Andy V. Huynh\inst{1,2,5}\orcidlink{0009-0003-2892-706X} \and
Lauren E. Gillespie\inst{1,2,3,4}\orcidlink{0000-0003-2496-8035} \and
Jael Lopez-Saucedo\inst{2}\and
Claire Tang\inst{2}\and
Rohan Sikand\inst{2}\and
Moisés Expósito-Alonso\inst{2,3,4,5}\orcidlink{0000-0001-5711-0700}}

\authorrunning{A. Huynh, L. E. Gillespie, et al.}

\institute{
Co-first authors\and
    Stanford University, Stanford CA 94305, USA \and
    Carnegie Science, Washington DC 20015, USA \and
    University of California, Berkeley, Berkeley CA 94720, USA \and
    Howard Hughes Medical Institute,  Chevy Chase MD  20815, USA \\
\email{\{avhuynh,gillespl\}@cs.stanford.edu} }

% \vspace{-0.7cm}
\maketitle

\setcounter{table}{7}
\section{Nature Multi-View Dataset}
Here we describe the Nature Multi-View (NMV) dataset in specific detail.
\subsection{iNaturalist observations}\label{inat_obs}
To build the NMV dataset, we first downloaded the latest iNaturalist observations from the iNaturalist open data Amazon S3 Bucket as of September 27th, 2023, which totaled 119,687,551 observations. To curate these observations, we then in order:
\begin{itemize}
    \item Removed all casual-grade observations (which denote captive or cultivated individuals, such as cultivated plants in gardens), leaving only observations that are research-grade or still in need of ID. %This removed ~5\% of the dataset, leaving 113,866,211 observations.
    \item Filtered to observations with a high positional accuracy. Most iNaturalist observations come with a provided latitude and longitude, indicating where the observation was acquired on Earth. When an observation is taken with a smartphone (as the majority of observations are), these coordinates are provided by the smartphone's GPS system, which can estimate the radius of uncertainty of the measured geographic location. To ensure that the proper remote sensing image is co-registered with the ground-level observation, we removed any observation with an estimated positional accuracy radius of $>$ 120m, ensuring that every ground-level photo should geographically be located somewhere within the corresponding 153x153m-wide remote sensing image. %This removed ~52\% of the remaining observations, leaving 59,192,109 observations.
    \item Focused on plants. We chose to focus specifically on plants instead of animals, as they do not move within their lifetime and thus can be co-registered with remote sensing imagery from a broader time horizon.% Filtering to kingdom Plantae removed ~43.5\% of the remaining observations, leaving 25,740,163 observations.
    \item Kept terrestrial vascular plant observations. Non-vascular plants, including mosses, liverworts, hornworts, green algae, and red algae are also members of kingdom Plantae along with the more common vascular plants. However, these non-vascular plants are highly evolutionarily distinct from the rest of the plant kingdom and tend to be either unicellular, aquatic-based, and relatively rare. We specifically only kept observations from phylum Tracheophyta. %They also only represent ~2.5\% of the remaining observations from kingdom Plantae, thus we removed them, leaving 25,076,655 observations. 
    \item Filtered to contemporary observations. Most iNaturalist observations are taken in the app and are from after the public release of the site, but since observations can in theory be manually uploaded from any timepoint, we only kept observations taken on or after January 1st, 2011. %This removed XX\% of the dataset, leaving XXX observations
    \item Keept only observations within the administrative bounds of California. We used the Global Administrative Area-defined boundaries of California to demarcate which observations were found within the state's bounds.
    \item Filtered to observations that were co-registered with valid bioclimatic data. We specifically co-registered all observations with  30 arc-second resolution WorldClim bioclimatic variables \cite{fick_worldclim_2017}.
    \item Finally, we re-classified all observations up to the species level. Some observations are identified at a finer taxonomic resolution than the species level (e.g. subspecies, varieties), so to ensure consistency we joined these taxonomies up to the species level.

\end{itemize}
\subsection{Ground-level view images}\label{inat_img}

After curating the iNaturalist observations, we then linked each observation to the photos associated with that observation using the provided metadata. In our curated dataset, each observation on average has almost two ground-level images associated with it, often providing multiple separate ground-level views of the individual, such as the bark of a tree in one image and a picture of the canopy in another. We chose to include all of these ground-level observations in the NMV dataset, as they provide additional views and details for each observation. 

We downloaded all possible images from the S3 bucket, but not every observation from our original curated list had at least one available ground-level views to download, usually resulting from iNaturalist users revoking the necessary permissions for an observation's photos to be included in the iNaturalist Open Data repository. We subsequently excluded the few dozen observations whose ground-level photos were no longer available from the NMV dataset.

\subsection{Aerial view images}\label{nmv_naip}
After downloading the ground-level view images, we then co-registered each curated iNaturalist observation with a 256x256-pixel RGB-Infrared image from the 2018 acquisition of the National Aerial Imagery Program (NAIP) \cite{USDA_NAIP} using the provided coordinates. Briefly, for each observation, the NAIP GeoTiffs overlapping the coordinates of the observation were selected, then a 256x256-pixel cropped image was taken from the GeoTiff, with the crop centered at the pixel co-registered with the geographic coordinate of the observation. Given that NAIP imagery acquired in 2018 was taken at 60cm resolution, this generated aerial imagery capturing the approximately 153x153m area surrounding a given  observation. For a few hundred observations, 2018 NAIP imagery was not available (NAIP imagery is not publicly released around sensitive military sites such as Vandenberg Space Force Base). Like the ground-level view imagery, these observations were also filtered out of the NMV dataset.

\subsection{Data licenses, permissions, and reuse}
\label{sec:license}
All the images available in the public iNaturalist dataset---and therefore the ground-level view images in the NMV dataset---are distributed under one of the following creative commons licenses: CC0, CC-BY, CC-BY-SA, CC-BY-ND, CC-BY-NC, CC-BY-NC-SA, CC-BY-NC-ND. This mix of licenses means that the iNaturalist photos in the NMV dataset can be copied and distributed for non-commercial or research purposes, making them ideal for open-access model development. All iNaturalist images displayed in this publication are CC0, or public domain, but many images in the iNaturalist dataset require attribution to be given to the creator, do not allow commercial uses of the images, and/or do not allow adaptation of the images, only distribution and copying. License information and iNaturalist user IDs are provided in the NMV train, test, and validation datasets to ensure dataset users can know which license is relevant for what images and can give proper attribution. It is the responsibility of dataset users to ensure that their usage of iNaturalist images in the NMV dataset fall within the permissions granted under the respective creative commons licenses for those images. All NAIP imagery is public domain (CC0) and can be used and adapted for any commercial or non-commercial purpose without attribution, therefore there are no use or attribution requirements for the aerial view images in the NMV dataset.

\subsection{Train / test / validation splits} \label{inat_crossval}
We chose to use a spatial block holdout to generate train, test, and validation data splits. First we generated 0.1 x 0.1 degree boxes (using WGS 84 coordinate reference system) overlapping the outline of California, then randomly selected 12.5\% of these boxes for the test and validation splits. We then assigned NMV observations to one of the splits based on which block the observation fell within. Due to variation in the geographic distribution of iNaturalist observations across the state, the partitioned train, test, and validation observations came out to ~75\%, 11\%, and 14\% of the observations, respectively. 

For the test and validation observations, we filtered out any observation that was within 256m of an observation in the training split (corresponding to observations very close to the edge of the spatial blocks). We further filtered these observations to only keep research-grade observations and observations identified down to the species level. In general, we considered research-grade observations identified down to the species level as of sufficient quality to use as the ground-truth species label for downstream species classification and distribution modeling. For the train split of the dataset, we labeled these classification-quality observations as fully-labeled training observations. Finally, the test, validation, and fully-labeled training observations were further filtered to only include observations for species that are present at least once in all three splits.

\subsection{Fine-tuning training data ratios}\label{inat_splits}
For generating the fine-tuning training data percentages ($\lambda$), we aimed to choose percentages that mimic the density of biodiversity data in regions outside of California. To do so, for the 119,687,551 total observations available globally from iNaturalist used to generate the NMV dataset, we applied the curation process as outlined in S.M. \ref{inat_obs}, save the steps filtering to California and subsetting to areas with co-registered bioclimatic data. We then assigned each observation to its corresponding continent using the ESRI default world continents base map \cite{esri}. To calculate the density of observations per square kilometer, first we projected the continents into the pseudo-Mercator WGS 84 coordinate reference system, then calculated the area of each continent in square kilometers, and finally divided the number of filtered iNaturalist observations for each continent by its area (S.M. Table \ref{tbl:nmv_density}). 

\begin{table}[h]
\centering
\small
\caption{\textbf{iNaturalist data density by continent}}
\label{tbl:nmv_density}
\begin{tabular}{lccc}
\toprule
Region & Area $_{km^2}$ & Observations &  Density $_{obs/km^2}$ \\
\midrule
California & 649,64. & 1,974,623 & 3.0395 \\
Africa & 33,535,112 & 825,592 & 0.0246 \\

Asia & 114,529,491 & 798,609 & 0.00697 \\

Australia & 9,652,152 & 469,414 & 0.04863 \\

Europe & 35,087,458 & 3,470,485 & 0.0989 \\

North America & 111,314,433 & 8,763,094 & 0.0787 \\

Oceania & 658,167 & 327,042 & 0.4969 \\

South America & 20,684,393 & 204,811 & 0.009902 \\
\bottomrule
\end{tabular}
\end{table}

We provide four different fine-tuning training percentages ($\lambda$) from the set of fully-labeled training observations: 0.25\%, 1\%, 5\%, and 20\%. For hyperparameter tuning, we also include a 2.5\% percentage. For each $\lambda$, we assigned observations by randomly selecting $\lambda\%$ of the fully-labeled training observations to be included in the percentage. For fine-tuning models with a given $\lambda$, we further filtered the train, test, validation observations to only include species present in the intersection of the three sets.

\section{Model training details}\label{training}

For all models, hyperparameters were tuned using the 2.5\% fully-supervised fine-tune training percentage and the validation split of the NMV dataset. For fine-tuning, all models were trained for 25 epochs and accuracy was evaluated on the test set using early stopping on species Top-1 Accuracy.

\subsection{Supervised baselines}\label{full_sup}
For the supervised baselines, we trained a standard ResNet50 using similar hyperparameters to the NVIDIA ResNet v1.5 Imagenet training recipe \cite{nvidia_resnet}, namely using: 
\begin{itemize}
    \item a batch size of 256,
    \item stochastic gradient descent with momentum,
    \item a momentum of 0.875,
    \item a weight decay of $3.05$ x $10^{-5}$ applied to all non-bias and non-batch-normalized layers,
    \item cosine annealing learning rate scheduling,
    \item and cross entropy loss with 0.1 label smoothing as the objective.
\end{itemize}

%Many other popular self-supervised InfoNCE-based like SimCLR \cite{chen2020simple} and MoCo-v2 \cite{chen2020improved} utilize data augmentation as a critical component for generating the positive pair(s) for a given example. However, the standard suite of data augmentations for these approaches may not translate well to natural world imagery tasks where things such as color and texture can be integral for accurate identification of species \cite{vanhorn2021benchmarking}. Therefore, for CRISP and all fine-tuning, we do not apply any data augmentations to the ground-level and aerial images other than standard Imagenet / NAIP normalization, unless noted otherwise. 

\subsubsection{Ground-level view classifier (Imagenet)}
For the species recognition experiments, we trained an Imagenet-pretrained ResNet50 backbone with a learning rate of 0.256, a training recipe that has been shown to lead to strong performance on natural world classification tasks \cite{vanhorn2021benchmarking}. For data augmentation, each ground-level image was resized to 256x256 pixels with antialiasing and normalized with Imagenet means and standard deviations.

\subsubsection{Aerial view classifier (Random Init.)}
For the species distribution mapping experiments, we trained a randomly-initialized ResNet50 backbone baseline with a learning rate of 0.01. We originally tried using the same backbone and learning rate as the ground-level view classifier---namely using an Imagenet pre-trained ResNet50 backbone and a learning rate of  0.256---but saw worse validation set performance than training with a He-initialized random backbone and a learning rate of 0.01. With the randomly-initialized backbone, we also tested a higher learning rate of 0.256 but saw loss divergence. For data augmentation, each aerial image was resized to 256x256 pixels with antialiasing and normalized with means and standard deviations calculated per-channel for all NAIP images across California from 2012 \cite{gillespie_image_2022}.

\subsubsection{Mixture of experts}
During fine-tuning, to consider the effect of training with both views at the same time, we utilize a previously published mixture of experts procedure that learns a weighted combination between linear projections of the ground-level and aerial view encoder embeddings \cite{beery2022auto}. For mixture-of-experts training, each ground-level image was resized to 256x256 pixels with antialiasing and normalized with Imagenet means and standard deviations, each aerial image was resized to 256x256 pixels with antialiasing and normalized with means and standard deviations calculated per-channel for all NAIP images across California from 2012 \cite{gillespie_image_2022}, and a learning rate of 0.256 was used.

\subsection{CRISP}
For our ContRastive Image-remote Sensing Pre-training (CRISP) approach, we use two ResNet50 encoders with a 512 embedding dimension size \cite{mai2023csp}. When pre-training on the NMV dataset, we modify the number of channels of the remote sensing encoder to match the corresponding four channel RGB-Infrared aerial view NAIP imagery. We use the OpenCLIP implementation of the CLIP loss and a standard temperature of 2.659 \cite{radford2021learning, cherti_reproducible_2022}. Similar to the Imagenet backbone baseline from S.M. \ref{full_sup}, as the optimizer for CRISP we use stochastic gradient descent with a momentum of 0.875, a weight decay of $3.05$ x $10^{-5}$ applied to all non-bias and non-batch-normalized layers, and a cosine annealing learning rate schedule. Given that the Imagenet-pretrained baseline showed improved performance starting with Imagenet pre-trained weights for classification from ground-level view imagery but not aerial view imagery (see S.M. \ref{full_sup}), we pre-trained all CRISP models with an Imagenet pre-trained ResNet50 backbone for the ground-level encoder and a randomly-initalized ResNet50 backbone for the remote sensing encoder. 

\subsubsection{Standard CRISP (CRISP)\label{crisp}}
Standard CRISP uses the standard CLIP objective with no modifications \cite{radford2021learning}. During pre-training, for the ground-level encoder we resized the images to 256x256 pixels with antialiasing and normalized with Imagenet means and standard deviations. For the remote sensing encoder, we resized the images to 256x256 pixels with antialiasing and normalized with means and standard deviations calculated per-channel for all NAIP images across California from 2012 \cite{gillespie_image_2022}. For hyperparameter tuning, we tested a low learning rate (0.0005), a medium learning rate (.01), and a high learning rate (0.256). We found the medium learning rate provided better fine-tuning performance for both the ground-level and aerial views on the validation set, while the highest learning rate consistently led to loss divergence early in training. For standard CRISP pre-training, we used a learning rate of 0.01, a batch size of 350, and pre-trained for 12 epochs. For species recognition and species distribution model fine-tuning, the ground-level and aerial encoders were separately fine-tuned for 25 epochs using a learning rate of 0.01, a batch size of 256 and the same training recipe  and data augmentations as described in S.M. \ref{full_sup}. For mixture-of-experts fine-tuning, the same procedure as used for the supervised baselines from S.M. \ref{full_sup} was applied \cite{beery2022auto}. Specifically, both encoders were fine-tuned simultaneously for 25 epochs using a learning rate of 0.01, a batch size of 256 and the same training recipe as described in S.M. \ref{full_sup}, using the same data augmentations as the single-view fine-tuning.

\subsubsection{CRISP with remote sensing data augmentation (CRISP-Aug)\label{crisp_aug}}
Given that in the NMV dataset, on average two ground-level images map to the same aerial image, we also tested if with the standard CLIP loss, random cropping of the aerial image would improve downstream model performance. Specifically, for the aerial images we applied a random 100x100 pixel crop followed by a horizontal and vertical flip applied randomly with 50\% probability, along with a randomly-selected rotation in 90-degree increments \cite{9140372}, and lastly image normalization with means and standard deviations calculated per-channel for all NAIP images across California from 2012 \cite{gillespie_image_2022}. For the ground-level encoder, as with standard CRISP, we resized the images to 256x256 pixels with antialiasing and normalized with Imagenet means and standard deviations. For CRISP pre-training with remote sensing imagery augmentations, we used a learning rate of 0.01, a batch size of 350, and was pre-trained for 12 epochs. For species recognition and species distribution model fine-tuning, the ground-level and aerial encoders were separately fine-tuned for 25 epochs using a learning rate of 0.01, a batch size of 256 and the same training recipe  and data augmentations as described in S.M. \ref{full_sup}. 

\subsubsection{Many-to-one CRISP (CRISP-M2o)\label{crisp_m20}}
As a second approach to addressing the many ground-level images to one aerial image problem, we modified the standard CLIP loss so that in a given batch, if two images were from the same geographic area, the cosine similarity between \emph{all} of co-located ground-level images and \emph{all} of the co-located aerial images would be considered as correct pairs. Specifically, we calculated the geographic distances between all observations in the batch using the Haversine distance and considered all observations that fell within 250 meters to be from the same geographic area. For many-to-one CRISP pre-training, the same data augmentations as standard CRISP were used, along with a learning rate of 0.01, a batch size of 350, and the model was pre-trained for 8 epochs. For species recognition and species distribution model fine-tuning, the ground-level and aerial encoders were separately fine-tuned for 25 epochs using a learning rate of 0.01, a batch size of 256 and the same training recipe and data augmentations as described in S.M. \ref{full_sup}. 

\subsubsection{Parameterized CRISP (CRISP-Par)\label{crisp_par}}
As a second approach to addressing the many ground-level images to one aerial image problem, we modified the standard CLIP loss to include a learned dynamic scaling factor that differentially weights the contributions of the cosine similarity between each of the views to the loss. We used a similar parameterization to the mixture of experts procedure from S.M. \ref{moe}
 \cite{beery2022auto}. For parameterized CRISP pre-training, the same data augmentations as standard CRISP were used, along with a learning rate of 0.01, a batch size of 350, and the model was pre-trained for 8 epochs. For species recognition and species distribution model fine-tuning, the ground-level and aerial encoders were separately fine-tuned for 25 epochs using a learning rate of 0.01, a batch size of 256 and the same training recipe  and data augmentations as described in S.M. \ref{full_sup}.

\subsection{Masked Autoencoder (MAE) contrastive baselines}\label{mae}

For the first set of contrastive baselines, we pre-trained masked auto-encoders (MAEs) on the NMV dataset,  along with comparing to MAEs pre-trained on other large-scale imagery datasets \cite{MaskedAutoencoders2021, cong2022satmae}. These encoders were trained with only one view at a time.

\subsubsection{MAE pre-trained on Imagenet (MAE-Img)}
We first compared downstream species recognition performance from ground-level NMV imagery for masked autoencoders pre-trained on Imagenet-1k, specifically a base-sized vision Transformer (vit base patch16) \cite{MaskedAutoencoders2021}. We end-to-end fine-tuned this Imagenet MAE pre-trained encoder for 25 epochs with a batch size of 256 and the same training recipe as described in S.M. \ref{full_sup}, and augmented each ground-level image by resizing to 256x256 pixels with antialiasing and normalizing with Imagenet means and standard deviations. We tried fine-tuning with three different learning rates on the 2.5\% split of the validation set, namely a low learning rate (0.002), a medium learning rate (0.01, the default learning rate used for CRISP fine-tuning), and a high learning rate (0.256, the default learning rate for Imagenet backbone fine-tuning). We saw the best performance with the low learning rate of 0.002, and subsequently fine-tuned all MAE pre-trained encoders with this learning rate.

\subsubsection{SatMAE pre-trained on Functional Map of the World(SatMAE-FMoW)}
We next compared downstream species distribution modeling performance from aerial view NMV imagery for masked autoencoders pre-trained on the Functional Map of the World (FMoW) dataset, specifically a large vision Transformer trained with non-temporal checkpoints \cite{cong2022satmae}. We end-to-end fine-tuned this FMoW SatMAE pre-trained encoder for 25 epochs with a batch size of 256, a learning rate of 0.002, the same training recipe as described in S.M. \ref{full_sup}, and augmented each aerial image by resizing to 256x256 pixels with antialiasing and normalizing with means and standard deviations calculated per-channel for all NAIP images across California from 2012 \cite{gillespie_image_2022}. 

\subsubsection{MAE pre-trained on ground-level view NMV imagery (MAE-NMV)}
Using ground-level imagery from the NMV dataset, we also pre-trained a base-sized vision Transformer (vit base patch16) for 100 epochs with a batch size of 800 and standard training procedures from \cite{MaskedAutoencoders2021}, namely a cropped input size of 224, a mask ratio of 0.75, a weight decay of 0.05, using normalized pixel loss, a starting learning rate of 0.001, and using 5 warmup epochs. Data augmentations consisted of a 50\% chance random horizontal flip, resizing to 256x256 pixels with antialiasing and normalizing with Imagenet means and standard deviations. During pre-training, we saw that pre-training loss plateaued early, so we ended pre-training early after only 48 epochs. For species recognition, we end-to-end fine-tuned this ground level view NMV MAE pre-trained encoder for 25 epochs with a batch size of 256, a learning rate of 0.002, the same training recipe as described in S.M. \ref{full_sup}, and augmented each ground-level image by resizing to 256x256 pixels with antialiasing and normalizing with Imagenet means and standard deviations.

\subsubsection{SatMAE pre-trained on aerial view NMV imagery (SatMAE-NMV)}
Using aerial imagery from the NMV dataset, we also pre-trained a base-sized vision Transformer encoder (vit base patch16) for 100 epochs with a batch size of 800 and standard training procedures from \cite{cong2022satmae}, namely a cropped input size of 224, a mask ratio of 0.75, a weight decay of 0.05, using normalized pixel loss, a starting learning rate of 0.0024, and using 5 warmup epochs. Data augmentations consisted of a 50\% chance random horizontal and vertical flip, a 25\% chance 90 degree rotation, and image resizing to 256x256 pixels with antialiasing and normalization with means and standard deviations calculated per-channel for all NAIP images across California from 2012 \cite{gillespie_image_2022}. During pre-training, similar to SauMoCo we again saw that pre-training loss plateaued quite early, so we ended pre-training early after only 9 epochs. For species distribution modeling, we end-to-end fine-tuned this aerial view NMV SatMAE pre-trained encoder for 25 epochs with a batch size of 256, a learning rate of 0.002, the same training recipe as described in S.M. \ref{full_sup}, and augmented each aerial image by resizing to 256x256 pixels with antialiasing and normalization with means and standard deviations calculated per-channel for all NAIP images across California from 2012 \cite{gillespie_image_2022}.

\subsection{MoCo contrastive baselines pre-trained on NMV}\label{moco}
For the second set of contrastive baselines, we separately pre-trained ground-level and aerial view encoders using Momentum Contrast (MoCo) V2 \cite{chen2020improved, 9140372} with default MoCo hyperparameters using a ResNet50 encoder and a batch size of 256 for 100 epochs each \cite{vanhorn2021benchmarking}. For fine-tuning, we mimic standard MoCo feature extraction and only retain the pre-trained queue encoder minus the fully-connected linear projection head, which we replace with a randomly-initialized classification head. Unlike previous studies, \cite{vanhorn2021benchmarking,chen2020improved}, we fully end-to-end fine-tune these MoCo encoders.
\subsubsection{Ground-level view MoCo (MoCo-v2)}
For the ground-level view, we pre-trained MoCo using the same data augmentations as MoCo-v2 \cite{chen2020improved} with a modified color jitter. We found that pre-training with the default brightness, contrast, saturation, and hue jitter values got stuck very early into pre-training, but halving these values (specifically, a max brightness jitter of 0.2, contrast of 0.2, saturation of 0.2, and hue of 0.05 applied with the default 80\% chance of jitter) led to improved pre-training performance. During pre-training, we saw that both the pre-training loss and queue classification accuracy plateaued early and the change in fine-tuning accuracy was negligible, so we ended pre-training early and fine-tuned the ground-level view encoder after only 35 pre-training epochs based on when the top 1 queue classification accuracy plateaued. For species recognition, we end-to-end fine-tuned this ground level view MoCo-v2 pre-trained encoder for 25 epochs with a batch size of 256, a learning rate of 0.256, the same training recipe as described in S.M. \ref{full_sup}, and augmented each ground-level image by resizing to 256x256 pixels with antialiasing and normalizing with Imagenet means and standard deviations.

\subsubsection{Aerial view MoCo (SauMoCo)}
For the aerial view, we pre-trained MoCo using spatially-explicit augments from SauMoCo \cite{9140372}. Specifically, for both the query and key images, for each observation we take a random 100x100-pixel crop from the larger 256x256-pixel aerial image (equivalent to using a neighborhood of 50 pixels \cite{jean2019tile2vec}), then randomly apply a horizontal and vertical flip with 50\% probability, along with a randomly-selected rotation in 90-degree increments \cite{9140372}, and finally normalized with means and standard deviations calculated per-channel for all NAIP images across California from 2012 \cite{gillespie_image_2022}. To more accurately compare to the CRISP pre-training strategy, we limit the spatial crops to only the aerial views present in the NMV dataset, as opposed to all NAIP tiles falling within the training spatial blocks \cite{jean2019tile2vec, 9140372}. During pre-training, we saw that while the pre-training loss continued to improve over time, the queue classification accuracy quickly peaked after 3 epochs. Fine-tuning on the species distribution modeling task with the 2.5\% train data ratio and evaluating on the validation set showed negligible performance differences between the epoch with the highest queue classification accuracy and later epochs. Therefore, we ended pre-training early and fine-tuned the aerial view encoder after just the 3rd epoch of pre-training, when Top-1 queue classification accuracy was highest. For species distribution modeling, we end-to-end fine-tuned this aerial view MoCo-v2 pre-trained encoder for 25 epochs with a batch size of 256, a learning rate of 0.01, the same training recipe as described in S.M. \ref{full_sup}, and augmented each aerial image by resizing to 256x256 pixels with antialiasing and normalizing with means and standard deviations calculated per-channel for all NAIP images across California from 2012 \cite{gillespie_image_2022}. 

\begin{table*}
% \vspace{-0.4cm}
\small
\centering
\caption{\textbf{Top-1 Eco-Accuracy on NMV Dataset.} We report accuracies for models trained on either aerial or ground level imagery.}
\label{tbl:t1_eco_accuracy}
\vspace{-0.2cm}
\begin{tabular}{lcccccccc}
\toprule
& \multicolumn{4}{c}{Ground Level} & \multicolumn{4}{c}{Aerial} \\
\cmidrule(lr){1-1}\cmidrule(lr){2-5} \cmidrule(lr){6-9} 
 Train \% ($\lambda$)  &    0.25\% &  1\% &  5\% &  20\% &  0.25\% & 1\% & 5\% & 20\% \\
 % \cmidrule(lr){2-4} \cmidrule(lr){5-7} 
 % Train City &     Cupertino &    Sioux Falls &     Charlottesville &   Cupertino &    Sioux Falls &     Charlottesville \\
\cmidrule(lr){1-1}\cmidrule(lr){2-5} \cmidrule(lr){6-9}
CRISP &21.03&	\textbf{35.55}&	46.03&	59.36&	1.87&	1.98&	1.88&	2.21 \\
CRISP-Aug & \textbf{21.30} &	35.13&	46.06&	58.18&	\textbf{2.54}&	2.22&	2.13&	3.86\\
CRISP-M2o & 20.29 &34.32 &46.27 &59.40 &2.12 &2.37 &2.17 &2.45\\
CRISP-Par &20.64 &34.92 &\textbf{47.31} &59.21 &2.23 &\textbf{2.66} &2.06 &2.49 \\
\cmidrule(lr){2-5} \cmidrule(lr){6-9} 
MAE-NMV &2.88&12.30&32.31&54.38 & - & - & - & -\\
MAE-Img & 1.83&15.30&42.43&\textbf{64.38} & - & - & - & -\\
SatMAE-NMV& - & - & - & - & 0.50&2.06&\textbf{3.20}&\textbf{3.88}\\
SatMAE-FMoW & - & - & - & - &1.02&1.85&1.83	& 2.59\\
MoCo-v2   & 12.95&	23.85&	42.68&	61.93 & - &- & - & - \\
SauMoCo   & -  & - & - & - & 1.83& 2.54 &2.83 &	3.51  \\
Imagenet   &11.84 &27.16&	46.89&	64.30 & - & - & - & - \\
Random Init.  & - & - & - & - & 0.37&	1.62&	2.38&	3.28 \\
\bottomrule
\end{tabular}
% \vspace{-1.7\baselineskip}
\end{table*}
 \section{Experiments}
We compare the CRISP pre-training approach on five tasks: species recognition, species distribution mapping, species classification with mixture of experts, crop type mapping, and tree genus identification.

\subsection{Metrics}\label{metss}
For species recognition, we report standard Top-K accuracy for $k = 1$ and $5$. For species distribution mapping, since many species are expected to be present in each aerial image, we report top-K accuracy for $k = 5,30$, as is commonly done for species distribution mapping tasks \cite{cole_geolifeclef_2020}. We also report Top-1 accuracies for species distribution mapping in S.M. Table \ref{tbl:t1_sdm_acc}. Since Top-K accuracy is highly skewed towards frequent classes in long-tailed datasets like NMV, we also report averaged within-class Top-K accuracy for $k = 1$ for species recognition and $k = 30$ for species distribution mapping. We refer to this as Top-K macro-accuracy (Top-K Macro Acc.), also known as averaged recall or class-averaged recall \cite{beery2022auto}. 

We also report Top-1 accuracy for species recognition (S.M. Table \ref{tbl:t1_spec_recog_far}) and Top-30 accuracy for species distribution mapping (Table \ref{tbl:sdm_far}) binned by species frequency into frequent, common, and rare categories \cite{beery2022auto}. Using the species frequencies as calculated from all examples in the fully-labeled train, test, and validation splits, we defined species as frequent if they have more than 700 examples in the combined datasets, as rare if they have fewer than 200 examples in the combined labeled datasets, and as common otherwise, for a total of 868 frequent species, 862 common species, and 1,216 rare species.

For crop type mapping, like the Tile2Vec paper we report the standard accuracy score for all downstream classifiers \cite{jean2019tile2vec}. For the mixture of experts analysis, we report Top-1 accuracy, macro-accuracy, and also Top-1 accuracy averaged across ecoregions instead of classes, which we refer to as Top-1 Eco. accuracy. For consistency, we report Top-1 Eco. accuracy for the other views of the NMV dataset in S.M. Table \ref{tbl:t1_eco_accuracy}.
% We also report Top-5 Accuracy, Top-5 Macro Accuracy, and Top-1 Accuracy broken down into frequent, common, and rare classes.

\begin{table*}
% \vspace{-0.5cm}
\centering
\small
\caption{\textbf{Species recognition broken down by species frequency.}}
\label{tbl:t1_spec_recog_far}
\vspace{-0.3cm}
\begin{adjustbox}{width=\textwidth}
\begin{tabular}{ccccccccccccc}
\toprule
& \multicolumn{4}{c}{Top-1-F} & \multicolumn{4}{c}{Top-1-C} & \multicolumn{4}{c}{Top-1-R} \\
\cmidrule(lr){1-1}\cmidrule(lr){2-5} \cmidrule(lr){6-9} \cmidrule(lr){10-13}
Train \% ($\lambda$)  & 0.25\% & 1\% & 5\% & 20\% & 0.25\% & 1\% & 5\% & 20\% & 0.25\% & 1\% & 5\% & 20\% \\
\cmidrule(lr){1-1}\cmidrule(lr){2-5} \cmidrule(lr){6-9} \cmidrule(lr){10-13} 
CRISP & \textbf{20.06} & 44.66 & 60.14 & 70.67 & 0.67 & 8.48 & 26.58 & 45.56 & 0.09 & 1.20 & 4.45 & 13.23 \\
CRISP-Aug & 19.50 & 43.64 & 59.82 & 70.46 & 0.73 & 8.21 & 26.82 & 44.10 & 0.21 & 0.91 & 4.45 & 13.53 \\
CRISP-M2o & 19.50&43.57&59.21&71.01&0.58&8.27&25.44&45.77&0.56&0.83&4.71&13.47\\
CRISP-Par & 19.92&\textbf{45.33}& \textbf{60.64} &71.08&0.57&\textbf{9.14}&27.75&45.36&0.09&1.27&4.97&14.07\\
\cmidrule(lr){2-5} \cmidrule(lr){6-9} \cmidrule(lr){10-13}
MAE-NMV &1.63&11.95&43.75&65.85&0.00&0.01&9.05&38.16&0.00&0.00&0.36&8.09 \\
MAE-Img & 1.09&15.42&55.02&73.84&0.00&	0.05&	21.73&\textbf{53.16}&0.00&0.00&2.15&18.69\\
MoCo-v2 & 13.37 & 29.65 & 53.52 & 72.54 & \textbf{2.96} & 7.08 & 25.07 & 48.81 & \textbf{2.51} & 2.10 & 5.85 & 19.47 \\
Imagenet & 12.27 & 33.77 & 59.51 & \textbf{75.18} & 2.80 & 8.18 & \textbf{28.03} & 51.52 & 1.68 & \textbf{2.62} & \textbf{6.58} & \textbf{20.23} \\

\bottomrule
\end{tabular}
\end{adjustbox}
% \vspace{-2.0\baselineskip}
\end{table*}

\subsection{Species recognition}
\label{spec_rec}

For the species recognition task, we trained and evaluated models using all ground-level view images per-observation from the NMV dataset, treating multiple ground-level images taken of the same iNaturalist observation as separate examples. To be consistent with the species distribution mapping results, we also report Top-1 accuracies broken down into frequent, common, and rare classes as outlined in Table S.M. \ref{tbl:t1_spec_recog_far}.

\subsection{Species distribution mapping}
\label{sdm_exp}
For the species distribution mapping task, we trained and evaluated models using each unique aerial view associated with each iNaturalist observation in the NMV dataset, meaning that for observations with more than one ground-level image, the associated aerial image was only considered once. To be consistent with the species recognition results, we also report these Top-1 accuracies and Top-5 in Table S.M. \ref{tbl:t1_sdm_acc}.

\begin{table*}
\small
\centering
\caption{\textbf{Top-1 and Top-5 species distribution mapping accuracies.}}
\label{tbl:t1_sdm_acc}
\begin{tabular}{lcccccccccccc}
\toprule
& \multicolumn{4}{c}{Top-1 Acc} & \multicolumn{4}{c}{Top-5 Acc} & \multicolumn{4}{c}{Top-1 Macro Acc} \\
\cmidrule(lr){1-1}\cmidrule(lr){2-5} \cmidrule(lr){6-9} \cmidrule(lr){10-13}
Train \% ($\lambda$)  & 0.25\% & 1\% & 5\% & 20\% & 0.25\% & 1\% & 5\% & 20\% & 0.25\% & 1\% & 5\% & 20\%  \\
%\cmidrule(c){2-5} \cmidrule(c){6-9} \cmidrule(c){10-13}
\cmidrule(lr){1-1}\cmidrule(lr){2-5} \cmidrule(lr){6-9} \cmidrule(lr){10-13} 
CRISP & 2.43 & 2.21 & 2.00 & 2.08 & 8.30 &	7.36 &	6.52 &	6.90 & \textbf{0.90} & 0.87 & \textbf{1.00} & \textbf{1.16} \\
CRISP-Aug & 1.99 & 2.26 & 2.31 & 3.35 & 7.28 &	7.36 &	8.02 &	11.60& \textbf{0.90} & 0.83 & 0.98 & 1.09 \\
CRISP-M2o & 2.39& 2.20& 1.99& 2.06& 8.53& 7.56& 6.45& 7.05& 0.88& 0.95& 0.94& 1.13\\
CRISP-Par & \textbf{2.71}& 2.36& 2.10& 1.97& \textbf{ 9.30}& 7.92& 6.73& 6.44& 0.88& \textbf{0.97}& 0.99& 1.15\\
SatMAE-NMV & 1.35& \textbf{2.93}& \textbf{3.86}& \textbf{4.16} & 4.81& 9.50& \textbf{13.37}& \textbf{14.76}& 0.11& 0.28& 0.65& \textbf{1.16} \\
SatMAE-FMoW & 1.72& 2.55& 2.47& 2.59& 7.34& 8.76&9.12&9.44& 	0.31& 0.45& 0.44	&0.84 \\
SauMoCo & 2.41 &2.87 & 3.41 & 3.28 & 8.30 & 	\textbf{9.96} &	12.13 &	11.83 &  0.41 & 0.56 & 0.68 & 0.87 \\
Random Init. & 1.17 & 2.28 & 3.06 & 3.22 & 4.98 &8.36&	10.90&	11.26& 0.11 & 0.29 & 0.49 & 0.80 \\
\bottomrule
\end{tabular}
\end{table*}
\subsection{Mixture of experts fine-grained classification}
\label{moe}
For the mixture of experts fine-grained classification task, we fine-tune with each ground level--aerial image pair, including all paired images for observations with more than one ground-level image. For comparison to the single-view models, we also report Top-5 accuracies as well in Table S.M. \ref{tbl:t5_moe}.

\begin{table*}
\small
\centering
\vspace{-0.4cm}

\caption{\textbf{Top-5 mixture of expert accuracies.}}
\label{tbl:t5_moe}
% \vspace{-0.2cm}
\begin{tabular}{lcccccccc}
\toprule
& \multicolumn{4}{c}{Top-5 Acc} & \multicolumn{4}{c}{Top-5 Macro Acc} \\
\cmidrule(lr){1-1}\cmidrule(lr){2-5} \cmidrule(lr){6-9}
Train \% ($\lambda$)  & 0.25\% & 1\% & 5\% & 20\% & 0.25\% & 1\% & 5\% & 20\% \\
\cmidrule(lr){1-1}\cmidrule(lr){2-5} \cmidrule(lr){6-9} 
CRISP & \textbf{25.54} &	\textbf{60.62} &	\textbf{80.08} &	88.33&	\textbf{8.03}&	27.93&	51.36&	66.067 \\
Supervised & 12.56 & 58.42 &	78.92&	\textbf{88.68}&	4.16& \textbf{32.19} & 	\textbf{52.43} & \textbf{70.75}\\
\bottomrule
\end{tabular}

% \vspace{-2.7\baselineskip}
\end{table*}

% \begin{table*}
% % \vspace{-0.5cm}
% \centering
% \small
% \caption{\textbf{Mixture of experts accuracy broken down by species frequency.}}
% \vspace{-0.3cm}
% \begin{adjustbox}{width=\textwidth}
% \begin{tabular}{ccccccccccccc}
% \toprule
% & \multicolumn{4}{c}{Top-1-F} & \multicolumn{4}{c}{Top-1-C} & \multicolumn{4}{c}{Top-1-R} \\
% \cmidrule(lr){2-5} \cmidrule(lr){6-9} \cmidrule(lr){10-13}
% \% Train Set & 0.25\% & 1\% & 5\% & 20\% & 0.25\% & 1\% & 5\% & 20\% & 0.25\% & 1\% & 5\% & 20\% \\
% \cmidrule(lr){2-5} \cmidrule(lr){6-9} \cmidrule(lr){10-13} 
% CRISP & \textbf{5.71} &	32.66&	\textbf{60.90} &	73.76&	\textbf{0.18} &	4.09&	27.18&	47.80&	0.13&	0.91&	5.12&	15.29 \\
% Imagenet & 2.31 &	\textbf{33.38} &	60.01&	\textbf{75.31} &	0.02&	\textbf{7.59} &	\textbf{28.93} &	\textbf{53.59} &	\textbf{0.19} &	\textbf{2.37} &	\textbf{7.38} &	\textbf{22.38} \\
% \bottomrule
% \end{tabular}
% \end{adjustbox}
% \vspace{-2.0\baselineskip}
% \end{table*}

\subsection{Crop type mapping}
\label{crop_mapping}
For crop type mapping, we use the same southern Central Valley example from 2016 in central California as outlined in the original Tile2Vec paper \cite{jean2019tile2vec}. Specifically, we use the same train / test / validation spatial blocks, excluding the northernmost row of boxes for our experiment. From the train blocks, we randomly sampled 10,000 locations that are at least 128m away from validation or test blocks. For each randomly sampled location, we cropped a 256x256 pixel NAIP image taken in 2016, centered at the given random location and normaled with means and standard deviations calculated per-channel for all NAIP images across California from 2012 \cite{gillespie_image_2022}. We then used each model as a frozen feature extractor, extracting features from the 512-dimension final embedding layer of each model. In the comparison, we also include the publicy-available, pre-trained TileNet model \cite{jean2019tile2vec}. For TileNet, we further cropped each image to 50x50 pixels to match the size used during pre-training. Finally, as a trivial baseline, we use random Gaussian noise as the predictor for the downstream classifiers, approximating the accuracy when predicting just the most common class. 

To label each random example, from the USDA's 2016 cropland data layer (CDL) \cite{cdl_2016} we cropped a 154x154m block  corresponding to the full extent of the extracted 100 x 100 pixel NAIP image (154m in total length given that each NAIP pixel is 60cm in ground resolution). We then took the largest crop type class present in this pixel by area as the ground-truth crop type label for the image. For evaluating TileNet, we further cropped the CDL block to 30x30m to match the 50 x 50 pixel size of its training images. Finally, using the same process detailed above, we randomly generated 1,000 examples from the test blocks and extracted features for each example.

To evaluate performance, we fit three downstream classifiers---random forest, logistic regression, and multilayer perceptron---using the extracted features from each model. For each downstream classifier, we use the scikit-learn implementations with all default parameters other than logistic regression which uses the saga solver. We fit each downstream classifier both with 10,000 training examples and a randomly-selected subset of 1,000 training examples. Each downstream classifier was run four times and the mean accuracy across runs was reported. 

\subsection{Auto Arborist tree genus identification}
\label{aa}

\subsubsection{Dataset pre-training details}

For tree genus identification with the Auto Arborist dataset \cite{beery2022auto}, for the CRISP and MoCo-v2 self-supervised pre-training setup, we pre-trained using all AA train set images from all cities and all genera in the dataset. We also used sampling with replacement, meaning that a given example could end up as an in-batch negative for a given batch of pre-training. As the aerial images are only RGB, are lacking an infrared channel, and are taken from a different type of sensor than the NAIP imagery used in the NMV dataset, we normalize ground-level and aerial image views with Imagenet means and standard deviations during pre-training.

\subsubsection{Model pre-training details}
Like the NMV dataset, for CRISP pre-training we trained for 12 epochs with a learning rate of 0.01, a batch size of 350, and Imagenet-like optimization as outlined in S.M. \ref{full_sup}. The only additional data transformation applied to images during pre-training was to crop all images to 256x256 pixels with antialiasing. For the ground-level view MoCo-v2 pre-training, we applied the same image augmentations as outlined in S.M. \ref{moco}. Like the NMV dataset we pre-trained for 100 epochs with the same defaults and augmentations as outlined in S.M. \ref{moco}. Unlike the NMV dataset, we saw queue classification accuracy plateau quite early after only 12 epochs of pre-training, so we again ended pre-training early and evaluated all downstream fine-tuned models with weights from this epoch. For SauMoCo aerial view MoCo-v2 pre-training, we applied the same image augmentations as outlined in S.M. \ref{moco}. Like the NMV dataset, we pre-trained for 100 epochs with the same defaults and augmentations as outlined in S.M. \ref{moco}. We again saw queue classification accuracy plateau quite early after only 11 epochs of pre-training, so we again ended pre-training early and evaluated all downstream fine-tuned models with weights from this epoch.

\subsubsection{Dataset fine-tuning details}
To evaluate the pre-trained models' downstream accuracy performance in low data label settings, we separately fine-tuned encoders on only one Auto Arborist city and one view at a time. The cities in the Auto Arborist dataset are partitioned into three regions (West, Central, East) and each region has a holdout city not seen during supervised training that is used to compare transfer performance across cities and the given region, specifically Santa Monica, CA for West (25,381 examples, 126 genera), Boulder, CO for Central (29,489 examples, 65 genera), and Pittsburgh, PA for East (23,382 examples, 79 genera). Because we are interested in comparing performance in low data and high diversity settings, for fine-tuning we selected one city from each of the three regions that had the highest genera diversity and had fewer observations than the holdout test city, namely Cupertino, CA for West (15,300 examples, 104 genera), Sioux Falls, SD for Central (13,277 examples, 137 genera), and Charlottesville, VA for East (1,571 examples, 56 genera). Our experiments consist of fine-tuning a given encoder and view on one of these cities, then testing the accuracy of the fine-tuned model on the holdout city for that region. For example, we fine-tune a ground-level encoder on ground-level imagery from Charlottesville, then test that fine-tuned model's performance on ground-level images from Pittsburgh. We calculated accuracy using all genera present in the test city. For data augmentation, we normalized both image views with Imagenet means and standard deviations during pre-training.

\subsubsection{Model fine-tuning details}
Similar to the NMV dataset, for 25  epochs we fine-tuned the CRISP aerial and ground view ResNet50 encoders separately, specifically with a batch size of 256, a learning rate of 0.01, and an Imagenet-like optimizer like S.M. \ref{full_sup}. For fine-tuning the MoCo-v2 pre-trained ground-level ResNet50 encoder, we fine-tuned for 25 epochs with a batch size of 256, a learning rate of 0.256, and an Imagenet-like optimizer like S.M. \ref{full_sup}.  For the SauMoCo pre-trained aerial ResNet50 encoder, we fine-tuned for 25 epochs with a batch size of 256, a learning rate of 0.01, and an Imagenet-like optimizer like S.M. \ref{full_sup}. For the supervised ground-level baseline, we started with an Imagenet-pretrained ResNet50 backbone and we fine-tuned for 25 epochs with a batch size of 256, a learning rate of 0.256, and an Imagenet-like optimizer like S.M. \ref{full_sup}. For the supervised aerial baseline, we started with a randomly initialized ResNet50 backbone and we trained for 25 epochs with a batch size of 256, a learning rate of 0.01, and an Imagenet-like optimizer like S.M. \ref{full_sup}. To be consistent with the results from the Auto Arborist paper, we also report Top-1 Macro Accuracy results in S.M. Table \ref{tbl:aa_t1_macro}.

\begin{table*}
\vspace{-0.4cm}
\small
\centering
\caption{\textbf{Top-1 Macro Accuracy on Auto Arborist Dataset.} We report accuracies for models trained on either aerial or ground level imagery.}
\label{tbl:aa_t1_macro}
\vspace{-0.2cm}
\begin{tabular}{lcccccc}
\toprule
& \multicolumn{3}{c}{Ground Level} & \multicolumn{3}{c}{Aerial} \\
\cmidrule(lr){1-1}\cmidrule(lr){2-4} \cmidrule(lr){5-7} 
 Test City &     Santa Monica &    Boulder &     Pittsburgh &     Santa Monica &    Boulder & Pittsburgh \\
 % \cmidrule(lr){2-4} \cmidrule(lr){5-7} 
 % Train City &     Cupertino &    Sioux Falls &     Charlottesville &   Cupertino &    Sioux Falls &     Charlottesville \\
\cmidrule(lr){1-1}\cmidrule(lr){2-4} \cmidrule(lr){5-7}
CRISP  & \textbf{10.21}&	\textbf{6.78}&	4.12&	\textbf{3.22} &	\textbf{2.74}&	\textbf{3.26} \\
MoCo-v2   & 4.87&	4.23&	\textbf{5.53} & - &- & - \\
SauMoCo   &  - & - & - & 2.76&	2.10&	1.74 \\
Imagenet   & 5.98&	3.32&	2.16 & - & - & - \\
Random Init.  & - & - & - & 1.76&	1.79&	1.94 \\
\bottomrule
\end{tabular}
% \vspace{-2.7\baselineskip}
\end{table*}
\subsection{CRISP embedding analysis}
\label{embeddings}
Here we describe the procedures behind the UMAP and K-Means embedding representation analyses.

\subsubsection{Generating the embeddings}
To generate embeddings for the UMAP visualization from our NMV-pretrained CRISP encoder, from
the fully labeled split of the NMV dataset (Table \ref{tbl:nmv} column 2) we randomly selected 4,000 aerial + ground-level pairs from each L3 ecoregion of the dataset (Fig. \ref{fig:f1}b) and extracted embeddings from both encoders individually for a total of 42,317 embeddings-per encoder (two ecoregions did not have 4,000 observations for embedding).

To generate embeddings for the many-to-one embedding analysis, we filtered all images from the fully labeled split of the NMV dataset (Table \ref{tbl:nmv} column 2) which came from iNaturalist observations with more than nine photos associated with the observation. This corresponds to ground level-aerial image pairs from the NMV training dataset where the aerial image has at least nine ground-level photos paired to it, representing an extreme case of the many ground-level to one aerial image scenario.

\begin{table}[!b]
{
% \vspace{-0.6cm}
\footnotesize
\centering
\label{tbl:clustering}
\caption{\textbf{Many to one embedding clustering analysis.} Clusters assigned with K-means++ for NMV observations with $\geq$ 9 ground-level images (97 observations, 1,020 images). Aug=Augmented, M2one=many-to-one, Comp=Completeness, Homog=Homogeneity, Adj=Adjusted MI=Mutual Information.}
\setlength\tabcolsep{2pt}
\begin{tabularx}{\linewidth}{c|c|c|c|c|c}
% {lllllllllllll}
% \toprule
Score & Comp & Homog. & V Measure & Adj. Rand & Adj. MI \\
\cmidrule(lr){2-6}

CRISP & 0.62062	&\textbf{0.54777}&	\textbf{0.58192}	&0.10349&0.27853\\
CRISP-Aug & 0.58566&0.52472&0.55352&0.07676&0.21819\\
CRISP-M2o & \textbf{0.62069}&0.54711&0.58158&\textbf{0.10774}&\textbf{0.27882} \\
Gaussian noise&0.46554&0.42644&0.44513&0.00069&0.00282\\
\cmidrule(lr){2-6}

\end{tabularx}
\label{tbl:clustering}
}
% \vspace{-.9\baselineskip}
\end{table}
\subsubsection{Aerial encoder UMAP projection}
For the aerial view CRISP pre-trained embeddings, we projected the embeddings into a 4-dimensional color space using UMAP with 4 components and cosine similarity as the distance metric. To remove outliers, we then calculated the modified z-score for each of the four color channels and removed any observation > 2 standard deviations from the mean for that channel, then re-scaled each channel to 0-1. We then plotted each color-projected embedding at the corresponding geographic location where the aerial photo was taken to generate the final image (Fig. \ref{fig:f3}a).
\subsubsection{Ground level encoder UMAP projection}
For the ground level view CRISP pre-trained embeddings, we again projected the embeddings into a 4-dimensional color space using UMAP with 4 components and cosine similarity as the distance metric. To remove outliers, we then calculated the modified z-score for each of the four color channels and removed any observation > 2 standard deviations from the mean for that channel, then re-scaled each channel to 0-1. We then calculated the average embedding value in color space for each genus by averaging the projected embeddings for all images of said genus per-color channel (Fig. \ref{fig:f3}b). 

We then mapped these average genus embedding colors to the taxonomic tree of genera in the NMV dataset. We plotted this circular taxonomy using the ETE phylogeny toolkit with the branch length equal to the genus' taxonomic hierarchy. In brief, circular phylogeny visualizations nest genera with a more similar taxonomic family history closer to each other on the circle, with the extending branches representing the higher taxonomic orders (family, order, kingdom, etc.) in the taxonomic tree. Even though genera are all visualized at the same leaf level in the taxonomic hierarchy, some leaves are visualized as longer than others for genera with fewer specialized clades in their taxonomy (e.g. gymnosperms versus Asteraceae), which means the lengths of the branches have no meaning, and only the relative distance relationship between leaves in the tree are meaningful.
\subsubsection{Many to one image clustering}
To compare how well the different CRISP approaches handle the many-to-one problem inherent in the NMV dataset (many ground-level images to one aerial image), 1,020 ground-level embeddings for the 97 observations in the NMV dataset with $\geq 9$ associated images were clustered using scikit-learn's K-Means++ clustering algorithm with all default parameters and 97 clusters (the number of unique observations) (S.M. Table \ref{tbl:clustering}). For a random baseline, a tensor of Gaussian random noise the same size as the CRISP embeddings was clustered as well. The goodness of how well photos from the same observation were clustered together was measured using five clustering metrics (homogeneity, completeness, V-measure, adjusted Rand index, and adjusted mutual information) with the observation labels used as the ground truth labels.
\newpage
\bibliographystyle{splncs04}
\bibliography{main}

\begin{thebibliography}{10}
\providecommand{\url}[1]{\texttt{#1}}
\providecommand{\urlprefix}{URL }
\providecommand{\doi}[1]{https://doi.org/#1}

\bibitem{inat}
www.inaturalist.org

\bibitem{cdl_2016}
of~Agriculture, U.D.: Usda national agricultural statistics service cropland data layer. \url{https://croplandcros.scinet.usda.gov/} (2016), accessed: 2022-10-16

\bibitem{DBLP:journals/corr/abs-2011-09980}
Ayush, K., Uzkent, B., Meng, C., Tanmay, K., Burke, M., Lobell, D.B., Ermon, S.: Geography-aware self-supervised learning. CoRR  \textbf{abs/2011.09980} (2020), \url{https://arxiv.org/abs/2011.09980}

\bibitem{Bastani_2023_ICCV}
Bastani, F., Wolters, P., Gupta, R., Ferdinando, J., Kembhavi, A.: Satlaspretrain: A large-scale dataset for remote sensing image understanding. In: Proceedings of the IEEE/CVF International Conference on Computer Vision (ICCV). pp. 16772--16782 (October 2023)

\bibitem{beery2022auto}
Beery, S., Wu, G., Edwards, T., Pavetic, F., Majewski, B., Mukherjee, S., Chan, S., Morgan, J., Rathod, V., Huang, J.: The auto arborist dataset: A large-scale benchmark for multiview urban forest monitoring under domain shift. In: Proceedings of the IEEE/CVF Conference on Computer Vision and Pattern Recognition. pp. 21294--21307 (2022)

\bibitem{botella2023overview}
Botella, C., Deneu, B., Gonzalez, D.M., Servajean, M., Larcher, T., Leblanc, C., Estopinan, J., Bonnet, P., Joly, A.: Overview of geolifeclef 2023: Species composition prediction with high spatial resolution at continental scale using remote sensing. Working Notes of CLEF  (2023)

\bibitem{Cai_2019_ICCV}
Cai, S., Guo, Y., Khan, S., Hu, J., Wen, G.: Ground-to-aerial image geo-localization with a hard exemplar reweighting triplet loss. In: Proceedings of the IEEE/CVF International Conference on Computer Vision (ICCV) (October 2019)

\bibitem{cepeda2023geoclip}
Cepeda, V.V., Nayak, G.K., Shah, M.: Geoclip: Clip-inspired alignment between locations and images for effective worldwide geo-localization. arXiv preprint arXiv:2309.16020  (2023)

\bibitem{chen2020improved}
Chen, X., Fan, H., Girshick, R., He, K.: Improved baselines with momentum contrastive learning (2020)

\bibitem{cherti_reproducible_2022}
Cherti, M., Beaumont, R., Wightman, R., Wortsman, M., Ilharco, G., Gordon, C., Schuhmann, C., Schmidt, L., Jitsev, J.: Reproducible scaling laws for contrastive language-image learning (Dec 2022). \doi{10.48550/arXiv.2212.07143}, \url{http://arxiv.org/abs/2212.07143}, arXiv:2212.07143 [cs]

\bibitem{cole_geolifeclef_2020}
Cole, E., Deneu, B., Lorieul, T., Servajean, M., Botella, C., Morris, D., Jojic, N., Bonnet, P., Joly, A.: The {GeoLifeCLEF} 2020 {Dataset}  (2020), \url{http://arxiv.org/abs/2004.04192}

\bibitem{Cole_2022_CVPR}
Cole, E., Yang, X., Wilber, K., Mac~Aodha, O., Belongie, S.: When does contrastive visual representation learning work? In: Proceedings of the IEEE/CVF Conference on Computer Vision and Pattern Recognition (CVPR). pp. 14755--14764 (June 2022)

\bibitem{cong2022satmae}
Cong, Y., Khanna, S., Meng, C., Liu, P., Rozi, E., He, Y., Burke, M., Lobell, D., Ermon, S.: Satmae: Pre-training transformers for temporal and multi-spectral satellite imagery. Advances in Neural Information Processing Systems  \textbf{35},  197--211 (2022)

\bibitem{deneu_participation_2020}
Deneu, B., Servajean, M., Bonnet, P., Munoz, F., Joly, A.: Participation of {LIRMM} / {Inria} to the {GeoLifeCLEF} 2020 challenge (Nov 2020), \url{https://hal.inria.fr/hal-02989084}

\bibitem{enquist_commonness_2019}
Enquist, B.J., Feng, X., Boyle, B., Maitner, B., Newman, E.A., Jørgensen, P.M., Roehrdanz, P.R., Thiers, B.M., Burger, J.R., Corlett, R.T., Couvreur, T.L., Dauby, G., Donoghue, J.C., Foden, W., Lovett, J.C., Marquet, P.A., Merow, C., Midgley, G., Morueta-Holme, N., Neves, D.M., Oliveira-Filho, A.T., Kraft, N.J., Park, D.S., Peet, R.K., Pillet, M., Serra-Diaz, J.M., Sandel, B., Schildhauer, M., Šímová, I., Violle, C., Wieringa, J.J., Wiser, S.K., Hannah, L., Svenning, J.C., McGill, B.J.: The commonness of rarity: {Global} and future distribution of rarity across land plants. Science Advances  \textbf{5}(11),  1--14 (2019). \doi{10.1126/sciadv.aaz0414}

\bibitem{esri}
ESRI: World continents base map. \url{https://hub.arcgis.com/datasets/esri::world-continents/explore} (2023), accessed: 2023-11-11

\bibitem{fick_worldclim_2017}
Fick, S.E., Hijmans, R.J.: {WorldClim} 2: new 1-km spatial resolution climate surfaces for global land areas. International Journal of Climatology  \textbf{37}(12),  4302--4315 (2017). \doi{https://doi.org/10.1002/joc.5086}, \url{https://rmets.onlinelibrary.wiley.com/doi/abs/10.1002/joc.5086}, \_eprint: https://rmets.onlinelibrary.wiley.com/doi/pdf/10.1002/joc.5086

\bibitem{garcin_plntnet-300k_2021}
Garcin, C., joly, a., Bonnet, P., Affouard, A., Lombardo, J.C., Chouet, M., Servajean, M., Lorieul, T., Salmon, J.: Pl@{ntNet}-{300K}: a plant image dataset with high label ambiguity and a long-tailed distribution. In: Vanschoren, J., Yeung, S. (eds.) Proceedings of the {Neural} {Information} {Processing} {Systems} {Track} on {Datasets} and {Benchmarks}. vol.~1. Curran (2021), \url{https://datasets-benchmarks-proceedings.neurips.cc/paper_files/paper/2021/file/7e7757b1e12abcb736ab9a754ffb617a-Paper-round2.pdf}

\bibitem{gillespie_image_2022}
Gillespie, L.E., Ruffley, M., Exposito-Alonso, M.: Deep learning models map rapid plant species changes from citizen science and remote sensing data. Proceedings of the National Academy of Sciences  \textbf{121}(37),  e2318296121 (2024). \doi{10.1073/pnas.2318296121}, \url{https://www.pnas.org/doi/abs/10.1073/pnas.2318296121}

\bibitem{Haas_2024_CVPR}
Haas, L., Skreta, M., Alberti, S., Finn, C.: Pigeon: Predicting image geolocations. In: Proceedings of the IEEE/CVF Conference on Computer Vision and Pattern Recognition (CVPR). pp. 12893--12902 (June 2024)

\bibitem{MaskedAutoencoders2021}
He, K., Chen, X., Xie, S., Li, Y., Doll{\'a}r, P., Girshick, R.: Masked autoencoders are scalable vision learners. arXiv:2111.06377  (2021)

\bibitem{DBLP:journals/corr/HeZRS15}
He, K., Zhang, X., Ren, S., Sun, J.: Deep residual learning for image recognition. CoRR  \textbf{abs/1512.03385} (2015), \url{http://arxiv.org/abs/1512.03385}

\bibitem{vanhorn2021benchmarking}
Horn, G.V., Cole, E., Beery, S., Wilber, K., Belongie, S., Aodha, O.M.: Benchmarking representation learning for natural world image collections (2021)

\bibitem{vanhorn2017devil}
Horn, G.V., Perona, P.: The devil is in the tails: Fine-grained classification in the wild (2017)

\bibitem{hu2018cvm}
Hu, S., Feng, M., Nguyen, R.M., Lee, G.H.: Cvm-net: Cross-view matching network for image-based ground-to-aerial geo-localization. In: Proceedings of the IEEE Conference on Computer Vision and Pattern Recognition. pp. 7258--7267 (2018)

\bibitem{jean2019tile2vec}
Jean, N., Wang, S., Samar, A., Azzari, G., Lobell, D., Ermon, S.: Tile2vec: Unsupervised representation learning for spatially distributed data. In: Proceedings of the AAAI Conference on Artificial Intelligence. vol.~33, pp. 3967--3974 (2019)

\bibitem{9140372}
Kang, J., Fernandez-Beltran, R., Duan, P., Liu, S., Plaza, A.J.: Deep unsupervised embedding for remotely sensed images based on spatially augmented momentum contrast. IEEE Transactions on Geoscience and Remote Sensing  \textbf{59}(3),  2598--2610 (2021). \doi{10.1109/TGRS.2020.3007029}

\bibitem{lacoste_geo-bench_2023}
Lacoste, A., Lehmann, N., Rodriguez, P., Sherwin, E.D., Kerner, H., Lütjens, B., Irvin, J.A., Dao, D., Alemohammad, H., Drouin, A., Gunturkun, M., Huang, G., Vazquez, D., Newman, D., Bengio, Y., Ermon, S., Zhu, X.X.: {GEO}-{Bench}: {Toward} {Foundation} {Models} for {Earth} {Monitoring} (Jun 2023). \doi{10.48550/arXiv.2306.03831}, \url{http://arxiv.org/abs/2306.03831}, arXiv:2306.03831 [cs]

\bibitem{li2021geographical}
Li, W., Chen, K., Chen, H., Shi, Z.: Geographical knowledge-driven representation learning for remote sensing images. IEEE Transactions on Geoscience and Remote Sensing  \textbf{60},  1--16 (2021)

\bibitem{Liu_2019_CVPR}
Liu, L., Li, H.: Lending orientation to neural networks for cross-view geo-localization. In: Proceedings of the IEEE/CVF Conference on Computer Vision and Pattern Recognition (CVPR) (June 2019)

\bibitem{mai2023csp}
Mai, G., Lao, N., He, Y., Song, J., Ermon, S.: Csp: Self-supervised contrastive spatial pre-training for geospatial-visual representations (2023)

\bibitem{Mall_2023_CVPR}
Mall, U., Hariharan, B., Bala, K.: Change-aware sampling and contrastive learning for satellite images. In: Proceedings of the IEEE/CVF Conference on Computer Vision and Pattern Recognition (CVPR). pp. 5261--5270 (June 2023)

\bibitem{graft-24}
Mall, U., Phoo, C.P., Liu, M.K., Vondrick, C., Hariharan, B., Bala, K.: Remote sensing vision-language foundation models without annotations via ground remote alignment. In: ICLR (2024)

\bibitem{manas_seasonal_2021}
Mañas, O., Lacoste, A., Giró-i Nieto, X., Vazquez, D., Rodríguez, P.: Seasonal {Contrast}: {Unsupervised} {Pre}-{Training} {From} {Uncurated} {Remote} {Sensing} {Data}. pp. 9414--9423 (2021), \url{https://openaccess.thecvf.com/content/ICCV2021/html/Manas_Seasonal_Contrast_Unsupervised_Pre-Training_From_Uncurated_Remote_Sensing_Data_ICCV_2021_paper.html}

\bibitem{nvidia_resnet}
NVIDIA: Resnet v1.5 for pytorch. \url{https://catalog.ngc.nvidia.com/orgs/nvidia/resources/resnet_50_v1_5_for_pytorch} (2023), accessed: 2023-11-24

\bibitem{oord_representation_2019}
Oord, A.v.d., Li, Y., Vinyals, O.: Representation {Learning} with {Contrastive} {Predictive} {Coding}. arXiv:1807.03748 [cs, stat]  (Jan 2019), \url{http://arxiv.org/abs/1807.03748}, zSCC: NoCitationData[s0] arXiv: 1807.03748

\bibitem{ouaknine_openforest_2023}
Ouaknine, A., Kattenborn, T., Laliberté, E., Rolnick, D.: {OpenForest}: {A} data catalogue for machine learning in forest monitoring (Nov 2023). \doi{10.48550/arXiv.2311.00277}, \url{http://arxiv.org/abs/2311.00277}, arXiv:2311.00277 [cs]

\bibitem{Pantazis_2021_ICCV}
Pantazis, O., Brostow, G.J., Jones, K.E., Mac~Aodha, O.: Focus on the positives: Self-supervised learning for biodiversity monitoring. In: Proceedings of the IEEE/CVF International Conference on Computer Vision (ICCV). pp. 10583--10592 (October 2021)

\bibitem{radford2021learning}
Radford, A., Kim, J.W., Hallacy, C., Ramesh, A., Goh, G., Agarwal, S., Sastry, G., Askell, A., Mishkin, P., Clark, J., Krueger, G., Sutskever, I.: Learning transferable visual models from natural language supervision (2021)

\bibitem{randin_monitoring_2020}
Randin, C.F., Ashcroft, M.B., Bolliger, J., Cavender-Bares, J., Coops, N.C., Dullinger, S., Dirnböck, T., Eckert, S., Ellis, E., Fernández, N., Giuliani, G., Guisan, A., Jetz, W., Joost, S., Karger, D., Lembrechts, J., Lenoir, J., Luoto, M., Morin, X., Price, B., Rocchini, D., Schaepman, M., Schmid, B., Verburg, P., Wilson, A., Woodcock, P., Yoccoz, N., Payne, D.: Monitoring biodiversity in the {Anthropocene} using remote sensing in species distribution models. Remote Sensing of Environment  \textbf{239},  111626 (Mar 2020). \doi{10.1016/j.rse.2019.111626}, \url{https://www.sciencedirect.com/science/article/pii/S0034425719306467}

\bibitem{sagawa2022extending}
Sagawa, S., Koh, P.W., Lee, T., Gao, I., Xie, S.M., Shen, K., Kumar, A., Hu, W., Yasunaga, M., Marklund, H., Beery, S., David, E., Stavness, I., Guo, W., Leskovec, J., Saenko, K., Hashimoto, T., Levine, S., Finn, C., Liang, P.: Extending the wilds benchmark for unsupervised adaptation (2022)

\bibitem{sastry2024birdsat}
Sastry, S., Khanal, S., Dhakal, A., Huang, D., Jacobs, N.: Birdsat: Cross-view contrastive masked autoencoders for bird species classification and mapping. In: Proceedings of the IEEE/CVF Winter Conference on Applications of Computer Vision. pp. 7136--7145 (2024)

\bibitem{shi2019spatial}
Shi, Y., Liu, L., Yu, X., Li, H.: Spatial-aware feature aggregation for image based cross-view geo-localization. Advances in Neural Information Processing Systems  \textbf{32} (2019)

\bibitem{Shugaev_2024_WACV}
Shugaev, M., Semenov, I., Ashley, K., Klaczynski, M., Cuntoor, N., Lee, M.W., Jacobs, N.: Arcgeo: Localizing limited field-of-view images using cross-view matching. In: Proceedings of the IEEE/CVF Winter Conference on Applications of Computer Vision (WACV). pp. 209--218 (January 2024)

\bibitem{DBLP:journals/corr/abs-2108-05094}
Swope, A.M., Rudelis, X.H., Story, K.T.: Representation learning for remote sensing: An unsupervised sensor fusion approach. CoRR  \textbf{abs/2108.05094} (2021), \url{https://arxiv.org/abs/2108.05094}

\bibitem{teng2023satbird}
Teng, M., Elmustafa, A., Akera, B., Bengio, Y., Abdelwahed, H.R., Larochelle, H., Rolnick, D.: Satbird: Bird species distribution modeling with remote sensing and citizen science data. arXiv preprint arXiv:2311.00936  (2023)

\bibitem{tuia2022perspectives}
Tuia, D., Kellenberger, B., Beery, S., Costelloe, B.R., Zuffi, S., Risse, B., Mathis, A., Mathis, M.W., van Langevelde, F., Burghardt, T., et~al.: Perspectives in machine learning for wildlife conservation. Nature communications  \textbf{13}(1), ~792 (2022)

\bibitem{USDA_NAIP}
{United States Department of Agriculture}: {NAIP Imagery}. \url{https://naip-usdaonline.hub.arcgis.com/} (2023)

\bibitem{Van_Horn_2021_CVPR}
Van~Horn, G., Cole, E., Beery, S., Wilber, K., Belongie, S., Mac~Aodha, O.: Benchmarking representation learning for natural world image collections. In: Proceedings of the IEEE/CVF Conference on Computer Vision and Pattern Recognition (CVPR). pp. 12884--12893 (June 2021)

\bibitem{Horn_2018_CVPR}
Van~Horn, G., Mac~Aodha, O., Song, Y., Cui, Y., Sun, C., Shepard, A., Adam, H., Perona, P., Belongie, S.: The inaturalist species classification and detection dataset. In: Proceedings of the IEEE Conference on Computer Vision and Pattern Recognition (CVPR) (June 2018)

\bibitem{vo2016localizing}
Vo, N.N., Hays, J.: Localizing and orienting street views using overhead imagery. In: Computer Vision--ECCV 2016: 14th European Conference, Amsterdam, The Netherlands, October 11--14, 2016, Proceedings, Part I 14. pp. 494--509. Springer (2016)

\bibitem{workman2015wide}
Workman, S., Souvenir, R., Jacobs, N.: Wide-area image geolocalization with aerial reference imagery. In: Proceedings of the IEEE International Conference on Computer Vision. pp. 3961--3969 (2015)

\bibitem{10.1145/3394171.3413896}
Zheng, Z., Wei, Y., Yang, Y.: University-1652: A multi-view multi-source benchmark for drone-based geo-localization. In: Proceedings of the 28th ACM International Conference on Multimedia. p. 1395–1403. MM '20, Association for Computing Machinery, New York, NY, USA (2020). \doi{10.1145/3394171.3413896}, \url{https://doi.org/10.1145/3394171.3413896}

\bibitem{Zhu_2021_CVPR}
Zhu, S., Yang, T., Chen, C.: Vigor: Cross-view image geo-localization beyond one-to-one retrieval. In: Proceedings of the IEEE/CVF Conference on Computer Vision and Pattern Recognition (CVPR). pp. 3640--3649 (June 2021)

\end{thebibliography}
\end{document}